%% file: arxiv_main.tex
\documentclass[oneside,11pt]{article} 

\usepackage{algorithm,algorithmic}
\usepackage{amssymb,amsmath,wrapfig,amsfonts,amsthm}
\usepackage{xcolor}
\usepackage{graphicx}
\usepackage{booktabs}
\usepackage{multirow}
\usepackage{xspace}
\numberwithin{equation}{section}

\newtheorem{theorem}{Theorem}
\newtheorem{lemma}[theorem]{Lemma}

\newtheorem{corollary}[theorem]{Corollary}
\newtheorem{proposition}[theorem]{Proposition}
\newtheorem{definition}{Definition}
\newtheorem{assumption}{Assumption}

\theoremstyle{definition}
\newtheorem{remark}[theorem]{Remark}

\renewcommand{\bar}[1]{\overline{#1}}

\newcommand{\A}{\mathcal{A}}

\renewcommand{\P}{\mathcal{P}}
\renewcommand{\H}{\mathcal{H}}
\newcommand{\F}{\mathcal{F}}
\DeclareMathOperator{\E}{\mathbb{E}}
\newcommand{\R}{\mathbb{R}}

\newcommand{\poly}{\mathrm{poly}}
\newcommand{\polylog}{\mathrm{polylog}}
\newcommand{\maj}{\mathrm{Maj}}

\newcommand{\err}{\mathrm{err}}

\renewcommand{\(}{\left(}
\renewcommand{\)}{\right)}
\newcommand{\sign}[1]{\mathrm{sign}\(#1\)}
\newcommand{\D}{\mathcal{D}}
\newcommand{\X}{\mathcal{X}}
\newcommand{\Y}{\mathcal{Y}}

\newcommand{\PC}{\mathcal{P}}
\newcommand{\PL}{\mathcal{P}}

\newcommand{\tmin}{t_{\min}}
\newcommand{\tmax}{t_{\max}}
\newcommand{\assign}{\leftarrow}

\newcommand{\labeloverhead}{\Lambda_{L}}
\newcommand{\compareoverhead}{\Lambda_{C}}

\newcommand{\complabel}{\textsc{Compare-and-Label}\xspace}
\newcommand{\filter}{\textsc{Filter}\xspace}
\newcommand{\anti}{\textsc{Anti-Anti-Concentrate}\xspace}
\newcommand{\quicksort}{\textsc{Randomized Quicksort}\xspace}

\newcommand{\abs}[1]{\left\lvert #1 \right\rvert}

\graphicspath{{fig/}}

\usepackage{enumitem}

\newcommand{\citet}{\cite}
\newcommand{\citep}{\cite}

\title{Efficient PAC Learning from the Crowd with Pairwise Comparisons}
\author{
{Jie Shen}\\
\texttt{jie.shen@stevens.edu}\\
Stevens Institute of Technology
\and
{Shiwei Zeng}\\
\texttt{szeng4@stevens.edu}\\
Stevens Institute of Technology
}
\date{}
\begin{document}
\maketitle

\input{intro}
\input{related}

\input{setup}
\input{algorithm}

\input{conclusion}

\bibliography{../jshen_ref,../szeng_ref}
\bibliographystyle{alpha}

\clearpage
\appendix
\input{appendix}

\end{document}

%% file: intro.tex
\begin{abstract}
We study crowdsourced PAC learning of threshold functions, where the labels are gathered from a pool of annotators some of whom may behave adversarially. This is yet a challenging problem and until recently has computationally and query efficient PAC learning algorithm been established by Awasthi et al. (2017). In this paper, we show that by leveraging the more easily acquired pairwise comparison queries, it is possible to exponentially reduce the label complexity while retaining the overall query complexity and runtime. Our main algorithmic contributions are a comparison-equipped labeling scheme that can faithfully recover the true labels of a small set of instances, and a label-efficient filtering process that in conjunction with the small labeled set can reliably infer the true labels of a large instance set.
\end{abstract}

\section{Introduction} \label{sec:intro}

The recent years have witnessed an unprecedented growth of demands in annotating large-scale data sets via crowdsourcing. On the empirical side, crowdsourcing has been serving as a successful tool to harness the crowd wisdom for annotating images \citep{deng2009imageNet} and natural language \citep{callison2010speech}, evaluating and debugging machine learning models~\citep{chang2009reading}, performing hybrid intelligent system diagnosis~\citep{gomes2011crowd,tamuz2011adapt}, aiding our understanding in the conversation between human and intelligent systems \citep{doshivelez2017roadmap}, to name a few~\citep{vaughan2017making}.

Compared to the practical success, on the theoretical side, less is known about how to learn a hypothesis class with desired error rate from the noisy crowd annotations. This has motivated a flurry of recent studies on improving the quality of the annotations, such as pruning low-quality workers \cite{dekel2009vox}, adaptive task assignment~\citep{khetan2016achieve}, and self-correction~\citep{shah2016nooops}.

Orthogonal to these approaches, in this paper, we consider the generalization ability of learning algorithms when instances are annotated by the crowd. In the {\em standard probably approximately correct (PAC) learning model} of \citet{valiant1984theory}, it is assumed that there is a perfect hypothesis that incurs zero misclassification error (also known as the realizable setting), and the learner is always given correctly labeled data. In the context of crowdsourcing, the correct labels are not available but the learner has access to a large pool of workers and can give a task to multiple workers to aggregate the results, since some workers may behave adversarially~\citep{karger2011iterative,ho2013adaptive}. For example, the learner is allowed to query $k$ workers on any instance $x$ to gather a collection of noisy labels $Y(x) := \{y_1(x), \dots, y_k(x)\}$ and may, for example, take the majority vote as the final label for $x$. The goal is to find a hypothesis close to the perfect one without requesting too many crowd annotations.
The {\em crowdsourced PAC learning model} was recently put forward by \citet{awasthi2017efficient}, where they showed that any hypothesis class that can be efficiently learned in the standard PAC model can also be efficiently learned from the crowd even facing a high level of label noise. More importantly, it was shown that a carefully designed crowdsourcing algorithm collects an amount of labels that is only within a constant multiplicative factor of that needed in the standard PAC setting, which is called {\em a constant labeling overhead}. 

Generally speaking, the approach of \citet{awasthi2017efficient} crucially explores the wisdom of the crowd by adaptively gathering a manageable number of labels that can be aggregated into a set of correct labels, thus reducing the problem to the standard PAC learning setting. In this work, we consider the broad setting that the learner may additionally request comparison tags of any pair of instances to reduce the labeling cost, since in practice it is relatively easier to compare than to label. For example, in a recommendation application, an annotator may be asked which one of two movies she likes more; this is an easier task than to provide an exact rating, say, from 1 to 10~\cite{furnkranz2010prefer,park2015preference}. In general, we may consider that there is an underlying real-valued function $f^*(x)$ that indicates the user's preference to an instance $x$; the label query mathematically reads as ``is $f^*(x) > 0$?'' and the comparison query on two instances $(x, x')$ reads as ``is $f^*(x) > f^*(x')$?''. Such comparison-based model has been investigated in a variety of recent works in the context of PAC learning, aiming to mitigate the labeling cost or even break the label complexity lower bound known for the label-only setting; see, e.g.~\citet{kane2017active,hopkins2020noise} and the references therein for algorithms designed under the non-crowdsourcing model.

In light of the above, our goal is to develop an efficient crowdsourcing algorithm that: 1) returns a hypothesis with low error rate, namely, the PAC guarantee; 2) significantly mitigates the labeling cost through the more easily acquired pairwise comparisons, say, an $o(1)$ labeling overhead; and 3) achieves overall query complexity that is of the same order of that under the standard PAC learning model of \citet{valiant1984theory}, i.e. an overall $O(1)$ overhead.

Simultaneously achieving the three properties is highly nontrivial, since some natural approaches such as annotation followed by learning would lead to overhead blowing up with the sample size. In fact, we show that if the comparison overhead were not constrained, i.e. for each pair to be compared we can query as many workers as we want, then there would be a simple algorithm that achieves the first two properties, under the assumption that the majority of the crowd workers annotate according to the perfect hypothesis.\footnote{We phrase all the results by assuming some workers are perfect; it can be easily relaxed to being reliable; see Appendix~\ref{sec:app:massart}.}


\begin{theorem}[Theorem~\ref{thm:natural-approach}, informal]\label{thm:informal:natural}
Let $n$ be the sample size. Assume the majority of the crowd are perfect. Then any hypothesis class $\H$ can be learned from the crowd with $o(1)$ labeling overhead and $O(\log^2 n)$ comparison overhead as long as $\H$ is PAC-learnable in the standard setting.
\end{theorem}

One salient feature of the above result is that the labeling overhead becomes $o(1)$, whereas in the label-only crowdsourced PAC model of \citet{awasthi2017efficient} this bound is $O(1)$. However, we note that the comparison overhead grows with the sample size $n$. Even growing logarithmically, such phenomenon is counter-intuitive and undesirable in large-scale learning problems.

 


\vspace{-0.04in}
\subsection{Main results: an interleaving algorithm}
\vspace{-0.04in}


In this paper, we present a novel algorithm which well-controls the comparison overhead while retaining the label efficiency and PAC guarantee of Theorem~\ref{thm:informal:natural}. 

\begin{theorem}[Theorem~\ref{thm:main}, informal]\label{thm:main-informal}
There is an efficient crowsourcing algorithm that learns $\H$ with $o(1)$ labeling overhead and $O(1)$ comparison overhead provided that the majority of the crowd are perfect.
\end{theorem}

In stark contrast to the sample-size-varying comparison overhead of Theorem~\ref{thm:informal:natural}, the {constant overhead} we obtain here ensures that no matter what the size of data set is, the annotation cost  remains unchanged.


%
%

\subsection{Overview of techniques}


Our main algorithm is inspired by that of~\citet{awasthi2017efficient}, where the high-level idea is to utilize a celebrated boosting framework of \citet{schapire1990strength} for interleaving annotation and learning. It works as follows: first, train a good hypothesis $h_1$ on a small set of correctly labeled instances; second, identify and label a set of instances that $h_1$ may misclassify and train another good hypothesis $h_2$; last, feed $h_1$ and $h_2$ to the boosting framework to obtain a hypothesis with the desired error rate. Our main algorithmic contributions fall into new {\em label-efficient} processes to construct the data sets needed to train $h_1$ and $h_2$ by leveraging pairwise comparisons. We highlight the new techniques below and refer the readers to Section~\ref{sec:algorithm} for a detailed description.


%

\vspace{0.1in}
\noindent{\bfseries 1) Comparison-equipped labeling.} 
The first ingredient of our algorithm is a randomized sorting algorithm (Algorithm~\ref{alg:label}) which can sort all the instances in a given set $S_1$ with an $O(\abs{S_1}\cdot\log\abs{S_1})$ pairwise comparisons. It then suffices to perform binary search to find a ``threshold instance'' across which the labels shift, which consumes $O(\log\abs{S_1})$ label queries. We run this routine on a small set $S_1$ to train the hypothesis $h_1$ with desired error rate while keeping a low labeling and comparison overhead.

\vspace{0.1in}
\noindent{\bfseries 2) Label-efficient filtering with pairwise comparisons.}
The second ingredient, also the core component of our algorithm, is a label-efficient filtering process (Algorithm~\ref{alg:filter}) to identify instances that $h_1$ may make mistakes on. We construct an interval characterized by two ``support instances'' $x^-$ and $x^+$ with the properties that a) with high probability, $x^-$ is negative and $x^+$ is positive; b) instances outside the interval $[x^-, x^+]$ are likely to be good cases to test the performance of $h_1$, in the sense that their true labels can be {\em inferred} by combining a few comparison tags from the crowd and the label of $x^-$ or $x^+$. To this end, given an instance set $S$, we sub-sample a set $U$ and apply the comparison-equipped labeling to find proper support instances. The size of $U$ is carefully chosen, such that i) it is large enough so that the size of the interval $[x^-, x^+]$ is small and thus most instances in $S$ can be tested; and ii) it is not too large to blow up the overhead. We then show that the comparison complexity for identifying whether an instance is inside the interval, or outside that $h_1$ misclassifies, or outside that $h_1$ is correct on, is low. Since there are sufficient instances being tested, we are able to gather enough data to train $h_2$.


%

%% file: related.tex
\section{Related Works}\label{sec:related}

We discuss three research lines that are closely related to this work. In particular, we clarify the crucial connection and difference from noise-tolerant PAC learning, describe prominent crowd models and known results, as well as recent advances in PAC learning with comparisons.

\vspace{0.1in}
\noindent{\bfseries Noise-tolerant learning.} \
In real-world applications, the labels are often corrupted randomly or adversarially. To tackle such practical challenges, a variety of label noise models have been investigated in the literature, such as the random classification noise~\cite{angluin1987learning}, the Massart noise~\cite{sloan1992Corrigendum,massart2006risk}, the Tsybakov noise~\cite{tsybakov2004optimal}, and the agnostic/adversarial noise~\cite{haussler1992decision,kearns1992toward}. It is, however, known that without distributional or large-margin assumptions on the instances, learning even the fundamental class of linear threshold functions is computationally hard \cite{feldman2006new,guruswami2006hardness,diakonikolas2020near}. The only noise model that allows efficient and (almost) assumption-free algorithms is the random classification noise~\cite{blum1996polynomial}, yet it is recognized to be of little practical interest. Under distributional assumptions, a fruitful set of efficient PAC algorithms have been established, typically for the class of linear threshold functions; see, e.g. \citet{kalai2005agnostic,awasthi2017power,zhang2020efficient,diakonikolas2020learning,diakonikolas2020polynomial,shen2021attribute,shen2021power,shen2021sample,zhang2021improved} and the references therein. Notably, all these works approach the problem of learning linear threshold functions under the non-crowdsourcing scenario: for any given $x$, no matter how many times the learner queries its label, the obtained label is persistent even with noise. Orthogonal to these developed algorithms, we study the problem in the crowdsourcing setting and show that annotation aggregation enables a design of powerful PAC algorithms in the sense that our performance guarantee holds for general threshold functions {\em without} distributional assumptions on the instances (though we do need mild conditions on the crowd).

\vspace{0.1in}
\noindent{\bfseries Learning from the crowd.} \ 
Unlike traditional learning settings, a crowdsourced learner always has access to a large pool of workers who can provide annotations for a given instance. The noise in crowd mainly comes from the imperfectness of the workers. There are several types of crowd models considered in the literature. For example, one fundamental model assumes that a certain fraction of the workers are perfect, i.e. always labeling according to the true hypothesis, and the remaining could behave adversarially~\cite{karger2011iterative,awasthi2017efficient}, sometimes called the hammer-spammer model. Other works, such as \citet{welinder2010advance,ho2013adaptive,khetan2016achieve}, studied a more general crowd model where no perfect labeler exists, and the probability that a worker makes mistake depends on the given instance. The trouble of the latter model is that the annotators are too powerful and the algorithms had to either require golden labels \cite{ho2013adaptive}, or only showed weak guarantee that as the sample size $n$ goes to infinity, the misclassification error is small on average \cite{karger2011iterative,khetan2016achieve}; this could be vacuous since it is possible that the label of $\sqrt{n}$ instances can be entirely incorrect. To the best of our knowledge, the only known algorithm that offers PAC guarantee is due to \citet{awasthi2017efficient} under the fundamental crowd model (and a Massart-noise model).

\vspace{0.1in}\noindent
{\bfseries Learning with comparisons.} \
Pairwise comparisons have been widely applied in practical problems such as preference learning in recommender systems \cite{furnkranz2010prefer,park2015preference,xu2020simultaneous} and ranking \citep{JamiesonN11,heckel2019active,pananjady2017worst,shah2019feel}. More in line with this work is learning threshold functions with pairwise comparisons~\cite{kane2017active,xu2017noise,hopkins2020power,hopkins2020noise}, though these works are based on the non-crowdsourcing setting. The main finding is that it is possible to break the label complexity lower bound of label-only learning algorithms while still achieving PAC guarantees. We show similar phenomenon for crowdsourced PAC learning but will incorporate the comparison queries in a quite different way.

%% file: setup.tex
\vspace{-0.05in}
\section{Preliminaries} \label{sec:setup}


We study the problem of learning threshold functions, with access to both labels and pairwise comparison tags from the crowd. Let $\X$ be the instance space, and $\Y :=\{-1, 1\}$ be the label space. Let $\D$ be the joint distribution over $\X \times \Y$, and denote by $\D_{X}$ the marginal distribution over $\X$. Let $\F := \{ f: \X \rightarrow \R \}$ be the class of real-valued functions. The class of threshold functions is given by $\H := \{h: x \mapsto \sign{f(x)}, f \in \F\}$. For example, when $\F = \{f_w: x \mapsto w \cdot x\}$, the hypothesis class $\H$ is the class of halfspaces (also known as linear threshold functions). We focus on the classical realizable setting of \citet{valiant1984theory} where there exists a perfect hypothesis $h^* \in \H$ that incurs zero error, i.e. for any $(x, y)$ drawn from $\D$, $y = h^*(x)$. For any hypothesis $h: \X \rightarrow \Y$, its error rate is defined as $\err_{\D_X}(h) := \Pr_{x \sim \D_X}(h(x) \neq h^*(x))$. Let $f^* \in \F$ be the real-valued function associated with the perfect classifier $h^*$. For any pair of instances $(x, x') \in \X \times \X$, the underlying comparison function is thus given by $Z^*(x, x') = \sign{f^*(x) - f^*(x')}$. For the purpose of presentation transparency, for a given comparison function $Z(\cdot, \cdot)$, we will often write $x >_Z^{} x'$ and $x <_Z^{} x'$ in place of $Z(x, x') = 1$ and $Z(x, x') = -1$ respectively. When the function $Z$ is clear from the context, we omit the subscript and just write $x > x'$ or $x < x'$.

\vspace{0.1in}
\noindent
{\bfseries Standard PAC learning.} \
The terminology ``standard PAC learning'' is reserved for the setting of \citet{valiant1984theory}, that is, there exists a perfect hypothesis $h^*$ with zero error rate, and the learner is always given correctly labeled data $(x, h^*(x))$ where $x \sim \D_X$. We are interested in the classes $\H$ that are efficiently learnable in the standard PAC learning model.
\begin{assumption}\label{as:pac}
There exists an efficient algorithm $\A_\H$ which takes as input a target error rate $\epsilon \in (0, 1)$, confidence $\delta \in (0, 1)$, a set of $m_{\epsilon,\delta}$ correctly labeled instances $S$ randomly drawn from $\D$, and returns a hypothesis $h$ such that, with probability at least $1-\delta$, $\err_{\D_X}(h)\leq\epsilon$. We call $\A_\H$ a standard PAC learner.
\end{assumption}
For example, when $\H$ is the class of halfspaces, a linear program that fits the training data is also a standard PAC learner~\citep{maass1994fast}. Note that in Assumption~\ref{as:pac}, we did not impose distributional assumptions on $\D_X$.

The quantity $m_{\epsilon,\delta}$ in Assumption~\ref{as:pac} characterizes the query complexity of $\A_\H$.
In view of the classic result \cite{kearns1994intro}, it suffices to pick
\begin{equation}\label{eq:m}
m_{\epsilon, \delta} = K \cdot \frac{1}{\epsilon}\Big(d\log\frac{1}{\epsilon}+\log\frac{1}{\delta}\Big),
\end{equation}
where $d$ is the Vapnik–Chervonenkis dimension of the hypothesis class $\H$ and $K > 1$ is some absolute constant\footnote{For proper learning of Boolean-valued functions, there is a lower bound on the sample complexity that matches Eq.~\eqref{eq:m}; see the discussion in~\citet{hanneke2019website}.}. We reserve $m_{\epsilon, \delta}$ for the above expression. 
In our analysis, we will also need the following quantity
\begin{equation}\label{eq:m_eps}
n_{\epsilon,\delta} := K \cdot \frac{1}{\epsilon}\Big(d\log\frac{1}{\epsilon}+\Big(\frac{3}{\delta}\Big)^{\frac{1}{1000}} \Big),
\end{equation}
which will characterize the query complexity of a comparison-equipped algorithm. The additive term $(\frac{3}{\delta})^{\frac{1}{1000}}$ is  due to the fact that the failure probability of the sorting algorithm, specifically the randomized Quicksort algorithm we will use, only decays to zero in a rate inversely polynomial in the sample size. If there were sorting algorithms with failure probability decaying exponentially fast with the sample size, we would be able to choose $n_{\epsilon,\delta} = m_{\epsilon,\delta}$. Note also that there is nothing special on the exponent $\frac{1}{1000}$ of $\frac{1}{\delta}$:~it can be made to be a constant arbitrarily close to $0$ with the cost of increasing the comparison complexity with a constant multiplicative factor (see, e.g. Lemma~\ref{lem:quicksort}). 

\vspace{0.1in}
\noindent
{\bfseries Crowdsourced PAC learning.} \ 
Let $\PL$ be the uniform distribution over the pool of crowd workers who provide labels, and $\PC$ be the uniform distribution of those providing comparisons. In the context of crowdsourcing, the learner does not have access to the correct labels but has instances randomly drawn from $\D_X$, and it is allowed to query a worker $t \sim \PL$ on the label of an instance, or worker $t' \sim \PC$ on the comparison tag of two instances. For a given input (either a single instance or a pair), by querying multiple workers, it is possible to collect a set of different (noisy) annotations to aggregate (say, via majority voting). We say a crowdsourcing algorithm PAC learns $\H$ if for any given $\epsilon, \delta \in (0, 1)$, by drawing $m_S$ instances and collecting $m_L$ labels and $m_C$ comparison tags, it outputs a hypothesis ${h}: \X \rightarrow \Y$ such that with probability at least $1-\delta$, $\err_{\D_X}(h) \leq \epsilon$. We call $m_L$ the {\em label complexity}, and $m_C$ the {\em comparison complexity}. The query complexity refers to the sum of $m_L$ and $m_C$. The {\em labeling overhead} and {\em comparison overhead} are respectively defined as
\begin{equation}\label{eq:overheads-def}
\labeloverhead := \frac{m_L}{m_{\epsilon, \delta}}\quad \text{and}\quad \compareoverhead := \frac{m_C}{m_{\epsilon,\delta}},
\end{equation}
which characterize how the query complexity of a crowdsourced PAC learner compares to that of a standard PAC learner $\A_\H$. We say a crowdsourced PAC learner is query efficient if $\labeloverhead+\compareoverhead=O(1)$; if additionally $\labeloverhead = o(1)$, we say the learner is label efficient.


\vspace{0.1in}

\noindent{\bfseries Crowd model.} \
We consider the following fundamental model for the crowd: there exists a large pool of crowd workers, at least $\frac12+\alpha$ fraction of whom are perfect that always return correct labels (i.e. they label according to $h^*$), while the other $\frac12-\alpha$ fraction may perform adversarially. In other words, for a given instance, each time the learner queries a randomly chosen worker on its label, the obtained label is correct with probability at least $\frac12+\alpha$. Likewise, we assume that  $\frac12 + \beta$ fraction of the workers are perfect when providing the comparison tag. Throughout the paper, we assume that $\alpha\in(0,\frac12]$ and $\beta \in (0, \frac12]$, which ensures the correctness of the majority. It is worth mentioning that all of our analysis essentially holds for a much stronger noise model where the perfect workers can be replaced by {\em reliable} workers; we defer such extension to Appendix~\ref{sec:app:massart}.


Given a set of annotations $A = \{a_1, \dots, a_n\}$ (either the labels or comparison tags), we define $\maj(A)$ as the outcome of majority voting. Specifically, suppose $h_1$, $h_2$ and $h_3$ are three classifiers in $\H$. The function $\maj(h_1, h_2, h_3)$ maps any instance $x$ to a label $y$, which is the outcome of the majority vote of $h_1(x)$, $h_2(x)$ and $h_3(x)$. Let $Z_1, \dots, Z_t$ be $t$ comparison functions. We denote the set of the comparison tags given by them on a pair $(x, x')$ by $Z_{1:t}(x, x')$, namely, $Z_{1:t}(x, x') = \{ Z_1(x, x'), \dots, Z_t(x, x') \}$.

We will use $\tilde{O}(f)$ to denote $O(f \cdot \polylog(f))$, use $\tilde{\Omega}(f)$ to denote $\Omega(f / \polylog(f))$, and use $\tilde{\Theta}(f)$ to denote a quantity that is between them. The notation $O_{\delta}(f)$ means we treat the parameter $\delta$ as a constant, which will be useful to simplify our discussion on the overhead.

\begin{algorithm}[t]
\caption{$\textsc{Compare-and-Label}$}
\label{alg:label}
\begin{algorithmic}[1]

\REQUIRE A set of instances $S = \{x_i\}_{i=1}^n$, confidence $\delta$.
\ENSURE A labeled set $\bar{S}$.

\vspace*{2pt}
\STATE $k_1 \assign \frac{1}{2\beta^2} \cdot \log \frac{3006\abs{S} \log\abs{S}}{\delta}$ and $k_2 \assign \frac{1}{2\alpha^2} \cdot \log\frac{3\log\abs{S}}{\delta}$.

\STATE Apply \textsc{Randomized Quicksort} on $S$: for each pair $(x, x')$ being compared, query $k_1$ workers and take the majority. Let $\hat{S} = (\hat{x}_1, \hat{x}_2, \dots, \hat{x}_n)$ be the sorted list.
\STATE $\tmin \assign 1$, $\tmax \assign n$.

\WHILE{$\tmin < \tmax$}
\STATE $t \assign (\tmin + \tmax)/2$.

\STATE Query $k_2$ workers to obtain their labels on $\hat{x}_t$, and let $\hat{y}_t$ be the majority vote.

\STATE {\bfseries If} $\hat{y}_t = 1$ {\bfseries then} $\tmax \assign t - 1$; {\bfseries else} $\tmin \assign t+1$.

\ENDWHILE

\STATE For all $t' \geq t, \hat{y}_{t'} \assign 1$; for all $t' < t, \hat{y}_{t'} \assign -1$.

{\bfseries return} $\bar{S} \assign \{(\hat{x}_1, \hat{y}_1), (\hat{x}_2,\hat{y}_2), \dots, (\hat{x}_n, \hat{y}_n)\}$.

\end{algorithmic}
\end{algorithm}

%
%
%
%
%
%
%
%
%
%
%

\subsection{A natural approach and the limitation} \label{subsec:natural-approach}

In order to utilize the comparison queries, we consider a natural primitive of ``compare and label'' \cite{ailon2008efficient,xu2017noise}. The idea is to use pairwise comparisons along with the well-known \quicksort to sort all the instances. It then remains to perform binary search of a threshold instance across which labels shift from $-1$ to $1$; this step is label-efficient. 

The success of the above approach hinges on the correctness of the comparison tags and labels. To this end, for each input, the crowdsourcing algorithm may aggregate the annotations from multiple workers in order to ensure that the majority vote is correct with high probability. The details of \textsc{Compare-And-Label} are presented in Algorithm~\ref{alg:label}.

\begin{proposition}\label{prop:label-comp-complexity}
Consider the \complabel algorithm. If $\abs{S} \geq  (\frac{3}{\delta})^{1/1000}$, then with probability at least $1-\delta$, it correctly sorts and labels all the instances in $S$. The label complexity is $O\big(\frac{1}{\alpha^2}\cdot \log\abs{S}\cdot\log{\log\abs{S}} \big)$, and the comparison complexity is given by $O\big(\frac{1}{\beta^2}\cdot\abs{S}\cdot\log^2\abs{S}\big)$.
\end{proposition}


Equipped with the labeled set, it is straightforward to learn a classifier using the standard PAC learner $\A_\H$ as follows, where we recall that $n_{\epsilon,\delta}$ was defined in \eqref{eq:m_eps}.

\vspace{-0.03in}
\begin{quote}
{{\bfseries Natural Approach}}: Draw a set $S$ of size $n_{\epsilon,\delta}$ from $\D_X$. Let $\bar{S} \leftarrow \textsc{Compare-And-Label}(S)$. Return  $\A_\H(\bar{S}, \epsilon, \delta)$.
\end{quote}
\vspace{-0.03in}



We have the following performance guarantee.

\begin{theorem}[Natural approach]\label{thm:natural-approach}
With probability at least $1-\delta$, the natural approach runs in time $\poly(d,\frac{1}{\epsilon})$ and returns a classifier $h$ with error rate $\err_{\D_X}(h)\leq\epsilon$. The label complexity is $O\big(\frac{1}{\alpha^2} \cdot \log n_{\epsilon,\delta} \cdot \log {\log n_{\epsilon,\delta}}\big)$ and the comparison complexity is $O\big( \frac{1}{\beta^2} \cdot n_{\epsilon,\delta} \cdot \log^2 n_{\epsilon,\delta} \big)$. Moreover, the labeling overhead $\labeloverhead= \frac{1}{\alpha^2} \cdot \tilde{O}\big( \frac{\log(d/\epsilon)}{d/\epsilon} \big)$ and the comparison overhead $\compareoverhead= O_{\delta}\big( \frac{1}{\beta^2} \cdot \log^2 n_{\epsilon,\delta} \big)$.
\end{theorem}


\begin{remark}[Label complexity and labeling overhead]\label{rmk:label}
\citet{awasthi2017efficient} considered label-only crowdsourced PAC learning and obtained a label complexity bound of $O(m_{\sqrt{\epsilon},\delta} \log m_{\sqrt{\epsilon},\delta} + m_{\epsilon,\delta})$ and a labeling overhead $\labeloverhead' = O(\sqrt{\epsilon}\log\frac{m_{\epsilon,\delta}}{\delta}+1) = \tilde{O}_{\delta}\big( {\sqrt{\epsilon}\log(d/\epsilon)}+1\big)$ under the condition that $\alpha \geq \Omega(1)$. First, observe that our label complexity scales as $\log(d/\epsilon)$, while theirs is proportional to $d/{\epsilon}$ which is exponentially worse. Second, in terms of labeling overhead, our results show that it tends to zero as $\epsilon/d$ goes to zero, while their labeling overhead stays as a non-zero constant: this is precisely what we mean by $o(1)$ labeling overhead in Theorem~\ref{thm:informal:natural}. 
\end{remark}

However, we observe that the comparison overhead $\compareoverhead$ grows with the sample size. This is undesirable because when we demand a small error rate $\epsilon$, the sample size will increase and hence the comparison cost {per pair}. 

%% file: algorithm.tex
\section{Main Algorithm and Performance Guarantee} \label{sec:algorithm}

The natural approach separates learning and annotation by first labeling all the data and then training a classifier.
In this section, we present a more involved algorithm that interleaves learning and annotation to achieve constant comparison overhead. Inspired by the work of \citet{awasthi2017efficient}, we will use the celebrated boosting algorithm of \citet{schapire1990strength} as the starting point.

\begin{theorem}[Boosting]\label{thm:boost}
For any $p<\frac{1}{2}$ and distribution $\D_X$, consider three classifiers $h_1(x)$, $h_2(x)$, $h_3(x)$ satisfying the following. 1) $\err_{\D_X}(h_1) \leq p$; 2) $\err_{\D_2}(h_2) \leq p$ where $\D_2 := \frac{1}{2}\D_C + \frac{1}{2}\D_I$, $\D_C$ denotes the distribution $\D_X$ conditioned on $\{x: h_1(x)=h^*(x)\}$, and $\D_I$ denotes $\D_X$ conditioned on $\{x : h_1(x)\neq h^*(x)\}$; 3) $\err_{\D_3}(h_3) \leq p$ where $\D_3$ is $\D_X$ conditioned on $\{x: h_1(x)\neq h_2(x)\}$. Then $\err_{\D_X}(\maj(h_1, h_2, h_3)) \leq 3p^2-2p^3$.
\end{theorem}

There are two salient features that we will exploit from the theorem. First, the theorem implies that in order to learn a hypothesis with a target error rate $\epsilon \in (0, 1)$, it suffices to learn three weak hypotheses each of which has error rate less than $p = \frac{\sqrt{\epsilon}}{2}$. Second, the framework is fairly flexible in the sense that one can apply any algorithm to obtain the three nontrivial hypotheses, as long as it enjoys PAC guarantee in the distribution-independent regime. In view of Assumption~\ref{as:pac}, it suffices to gather $m_{\sqrt{\epsilon}/2, \delta/3}$ {\em correctly labeled} instances from $\D_X$, $\D_2$ and $\D_3$ respectively and invoke the standard PAC learner $\A_\H$.  Our main algorithm, Algorithm~\ref{alg:boost}, is designed in such a way that such an amount of high-quality labels are gathered without suffering an overwhelming query complexity (or, overhead). 

Our main theoretical results are as follows.

\begin{theorem}[Main results]\label{thm:main}
Consider Algorithm~\ref{alg:boost}. Assume $\beta \geq c_0$ for some absolute constant $c_0 \in (0, 1/2]$. For any $\epsilon, \delta \in (0, 1)$, with probability at least $1-\delta$, it runs in time $\poly(d,\frac{1}{\epsilon})$ and returns a classifier $\hat{h}$ such that $\err_{\D_X}(\hat{h})\leq\epsilon$. The label complexity is $\frac{1}{\alpha^2} \cdot \tilde{O}\big( \log\frac{d + \frac{1}{\delta}}{\epsilon} \big)$, and the comparison complexity is $\tilde{O}\big( \frac{d + (1/\delta)^{\frac{1}{1000}}}{\epsilon} \big)$. Moreover, the labeling overhead is $\frac{1}{\alpha^2} \cdot \frac{\epsilon}{d} \cdot \tilde{O}\big( \log\frac{d}{\epsilon} \big)$, and the comparison overhead is $O_{\delta}\big( \sqrt{\epsilon} \cdot \log^2\frac{d}{\epsilon} +1 \big)$.
\end{theorem}

\begin{remark}
The label complexity and labeling overhead are similar to what we obtained in Theorem~\ref{thm:natural-approach}, hence inheriting all the improvements over \citet{awasthi2017efficient} as we described in Remark~\ref{rmk:label}.
\end{remark}

\begin{remark}\label{rmk:constant_overhead}
A crucial message of the main theorem is that the comparison overhead can now be upper bounded by a constant; this occurs when $\epsilon$ is sufficiently small relative to the VC dimension $d$, i.e. $\epsilon \leq (\log d)^{-4}$. This is a mild condition since given a learning problem, the hypothesis class will be fixed and hence is the VC dimension $d$. Therefore, in the most interesting regime that $\epsilon$ is close to zero, i.e. $\epsilon \in (0, (\log d)^{-4})$, our algorithm PAC learns $\H$ with $o(\frac{1}{\alpha^2})$ labeling overhead and $O_{\delta}(1)$ comparison overhead. Here, we use $o(\frac{1}{\alpha^2})$ to emphasize that the labeling overhead may decay to zero as $\epsilon$ tends to zero. It is easy to see that if $\alpha$ is also lower bounded by some absolute constant, then the labeling overhead reads as $o(1)$.
\end{remark}

\begin{remark}
Our results demonstrate a useful tradeoff between the complexity of the two query types: though the overall query complexity is comparable to that of \citet{awasthi2017efficient}, our algorithm needs drastically fewer labels. This is especially useful for label-demanding applications such as what we introduced in Section~\ref{sec:intro}. One interesting question is the possibility to simultaneously reduce the label and overall query complexity. Unfortunately, there is evidence showing that this is in general impossible, though under the non-crowdsourcing setting; see e.g. Theorem 4.11 of \citet{kane2017active} and Theorem 11 of \citet{xu2017noise}.
\end{remark}

\begin{remark}\label{rmk:alternative_overhead}
Observe that in our definition of the overheads (see Eq.~\eqref{eq:overheads-def}), we compare the query complexity to that of a standard PAC learner that uses only labels. Hence, the notion of overheads highlights the savings of labels with pairwise comparisons. Another natural way to define the overheads is to compare the query complexity of a crowdsourcing algorithm to that of a non-crowdsourcing one which uses both labels and pairwise comparisons. This, however, appears subtle since it is unclear how the query complexity scales under the distribution-free setting. For example, combining Theorem~4.11 and Corollary~4.12 of \citet{kane2017active} results in a query complexity lower bound of $\Omega\big(d + \frac{1}{\epsilon}\big)$, yet its sharpness is open. We remark that if the query complexity were $\Theta\big(d + \frac{1}{\epsilon} \big)$, our main results would still follow for fixed VC dimension $d$. See Remark~\ref{rmk:overhead_analysis} in Appendix~\ref{sec:app:summary} for more details.
\end{remark}

In the following, we elaborate on the design of the main algorithm, along with the performance guarantees. To reduce clutter, we omit the confidence parameter $\delta$ in the discussion and  write $n_{\epsilon} := n_{\epsilon,\delta}$; it will be more convenient for the readers to further think of this quantity as being roughly $\Theta(\frac{1}{\epsilon})$. The precise form of $n_{\epsilon,\delta}$ (defined in Eq.~\eqref{eq:m_eps}) and the concrete choices of the confidence parameter will appear in the statements of all theorems and  algorithms. It will also be useful to keep in mind that feeding \complabel with a set of size $n_{\epsilon}$ would lead to comparison overhead growing with the sample size, but a set of size $n_{\sqrt{\epsilon}}$ would not since in this case the comparison complexity is $\tilde{O}(\frac{1}{\sqrt{\epsilon}})$ (see Theorem~\ref{thm:natural-approach}) because $m_{\epsilon, \delta} = \tilde{\Theta}(\frac{1}{\epsilon})$.


\begin{algorithm}[t]

\caption{Main Algorithm}
\label{alg:boost}

\begin{algorithmic}[1]

\REQUIRE Hypothesis class $\H$, target error rate $\epsilon$, confidence $\delta$, a standard PAC learner $\A_\H$.
\ENSURE A hypothesis $\hat{h}: \X \rightarrow \Y$.

\hspace{-12pt}{\textbf{Phase 1}}:

\STATE $\bar{S_1} \assign \complabel(S_1, \delta/6)$, for an instance set  $S_1$ of size $n_{\sqrt{\epsilon}/2,\delta/6}$ from $\D_{X}$.

\STATE $h_1'  \assign \A_\H(\bar{S_1}, \frac{\sqrt{\epsilon}}{2}, \frac{\delta}{6})$. \label{step:h_1'}

\STATE $h_1 \assign \anti(h_1', \frac{\sqrt{\epsilon}}{2}, \frac12)$.

\hspace{-12pt}{\textbf{Phase 2}}:

\STATE ${S_I} \assign \filter (S_2, h_1, \delta/12)$, for an instance set $S_2$ of size $\Theta(n_{\epsilon,\delta/12})$ drawn from $\D_X$.

\STATE $S_C \assign$ draw $\Theta(n_{\sqrt{\epsilon},\delta/12})$ instances from $\D_X$.

\STATE $\overline{S_{\text{All}}} \assign \complabel(S_I\cup S_C, \delta/12)$.

\STATE $\bar{W_I} \assign \{(x,y) \in \overline{S_{\text{All}}}: y\neq h_1(x) \}$,  $\bar{W_C} \assign \overline{S_{\text{All}}} \backslash \bar{W_I}$.

\STATE Draw a sample set $\bar{W}$ of size $n_{\sqrt{\epsilon}/2,\delta/12}$ from a distribution that equally weights $\bar{W_I}$ and $\bar{W_C}$.

\STATE $h_2 \assign \A_\H(\bar{W}, \frac{\sqrt{\epsilon}}{2}, \frac{\delta}{12})$.

\hspace{-12pt}{\textbf{Phase 3}}:

\STATE $\bar{S_3} \assign \complabel(S_3,\delta/6)$, for an instance set $S_3$ of size $n_{\sqrt{\epsilon}/2, \delta/6}$ from $\D_X$ conditioned on $h_1(x) \neq h_2(x)$.

\STATE $h_3 \assign \A_\H(\bar{S_3}, \frac{\sqrt{\epsilon}}{2}, \frac{\delta}{6})$.

{\bfseries return} {$\hat{h} \assign \maj(h_1, h_2, h_3)$.}

\end{algorithmic}

\end{algorithm}

\begin{algorithm}[t]

\caption{\anti}
\label{alg:anti}

\begin{algorithmic}[1]

\REQUIRE A classifier $h'$, a quantity $\eta \leq 1$ such that $\err_{D_X}(h') \leq \eta$, an absolute constant $c\leq 1$ with $\eta \leq c$.

\ENSURE A classifier $h$ with $\err_{D_X}(h) \in \big[c_1\eta, c_2\eta\big]$ where $c_1 = \min\big\{\frac12, \frac{1-c}{2-c}\big\}$, $c_2 = \max\big\{1, c+\frac12 \big\}$.

\STATE Pick bias $\lambda \in [0, 1]$ such that $(1-\lambda)(1-\frac12\eta) = \frac12\eta$.

\STATE Let $h$ be as follows: for any given $x$, independently toss a coin; {\bfseries if} outcome is HEADS (which happens with probability $\lambda$), {\bfseries then} $h(x) \assign h'(x)$, {\bfseries else} $h(x) \assign -h'(x)$.

{\bfseries return} $h$.
\end{algorithmic}	
\end{algorithm}

\subsection{Analysis of Phase 1}\label{subsec:phase_1}

In Phase 1, we aim to learn a nontrivial classifier $h_1$ with error rate at most $\frac{\sqrt{\epsilon}}{2}$ on $\D_X$. This can easily be fulfilled since $\D_X$ is available to the learner. In fact, this phase is very similar to the natural approach, except for the number of instances being used: here we draw $n_{\sqrt{\epsilon}/2} = \tilde{\Theta}(\frac{1}{\sqrt{\epsilon}})$ samples from $\D_X$ and feed them to \complabel, resulting in a nontrivial hypothesis $h_1'$ with $\err_{\D_X}(h_1') \leq \frac{\sqrt{\epsilon}}{2}$ and an $O(1)$ comparison overhead.

We then invoke \anti (Algorithm~\ref{alg:anti}) to prevent the performance of $h_1'$ from being too good. In particular, the new classifier $h_1$ is constructed in such a way that most of the time, it predicts as $h_1'$ does but with a small chance, it predicts as $-h_1'$. This would make sure that $\err_{\D_X}(h_1) = \Theta(\sqrt{\epsilon})$. While this step seems counter-intuitive, it is inherently important for our algorithm, especially for Phase~2. Indeed, the construction of $\D_2$ in Theorem~\ref{thm:boost} relies on the conditional distribution $\D_I$, consisting of instances from $\D_X$ that $h_1$ misclassifies. Since the probability density of $\D_I$ is exactly the error rate of $h_1$ on $\D_X$, in order to draw, say, $n$ instances from $\D_I$, we would have to sample at least $n/\err_{\D_X}(h_1)$ data from $\D_X$ in view of the Chernoff bound. Therefore, if the error rate of $h_1$ were not bounded from below, the number of samples to be drawn from $\D_X$ would be unbounded as well. One may question that if $h_1'$ is very good, say in reality we already have $\err_{\D_X}(h_1') \leq\epsilon$, then why not terminating Algorithm~\ref{alg:boost} and just return $h_1'$. The trouble here is that we cannot test its error rate, because the correct labels are not available. In fact, by the Chernoff bound, it is not hard to see that we would have to obtain $\tilde{\Theta}(1/\err_{\D_X}(h_1'))$ correctly labeled instances from $\D_X$ to guarantee that the empirical error rate of $h_1'$ equals $\Theta(\err_{\D_X}(h_1'))$, but a) the quantity $\err_{\D_X}(h_1')$ is unknown, and b) acquiring such amount of correct labels would lead to overwhelming overhead (once $\err_{\D_X}(h_1') = \epsilon$ this is exactly the amount needed by the natural approach).






We summarize the performance guarantee of Phase~1 below.
\begin{proposition}\label{prop:err-h1}
In Phase~1, with probability $1-\frac{\delta}{3}$, $\err_{\D_X}(h_1) \in \big[\frac{\sqrt{\epsilon}}{6}, \frac{\sqrt{\epsilon}}{2}\big]$. The label complexity is $O\big(\frac{1}{\alpha^2}\cdot\log n_{\sqrt{\epsilon},\delta}\cdot\log{\log n_{\sqrt{\epsilon},\delta}}\big)$
and the comparison complexity is $O\big(\frac{1}{\beta^2}\cdot n_{\sqrt{\epsilon},\delta} \cdot\log^2 n_{\sqrt{\epsilon},\delta}\big)$.
\end{proposition}

\begin{algorithm}[t]

\caption{\filter}
\label{alg:filter}

\begin{algorithmic}[1]

\REQUIRE An instance set $S$, classifier $h$ with $\err_{\D_X}(h) \in \big[\frac{\sqrt{\epsilon}}{6}, \frac{\sqrt{\epsilon}}{2}\big]$, confidence $\delta$.

\ENSURE A set $S_I$ whose instances are misclassified by $h$.

\STATE $b \assign \frac{4}{\sqrt{\epsilon}} \log\frac{16}{\delta} + (\frac{24}{\delta})^{1/1000}$.


\STATE Sample uniformly a subset $U \subset S$ of $b$ instances.

\STATE $\bar{U} \assign \complabel(U, \delta/3)$. Let $x^-$ be the rightmost instance of those labeled as $-1$, and $x^+$ be the leftmost instance of those labeled as $+1$.

\STATE $S_I \assign \emptyset$, $S_{\text{in}} \assign \emptyset$, $N \assign \frac{1}{\beta^2}\log\frac{1}{\epsilon}$.

\FOR{$x\in S\backslash U$}

\STATE $\text{ANS} \assign \text{YES}$.

\FOR{$t=1, \dots, N$}
\STATE Draw a worker $t \sim \PC$ to obtain the comparison tag $Z_t(x,x^-)$. {\bfseries If} $Z_t(x,x^-) = \{ x < x^-\}$, {\bfseries then} $Z_t(x,x^+) \assign \{x < x^+\}$, {\bfseries else} query  $Z_t(x,x^+)$. \label{step:draw}

\STATE \textbf{If} $t$ is even, \textbf{then} {\bfseries continue} to the next iteration.

{
\STATE \label{step:filter-outside} {Filtering: }\textbf{If} $\big[\maj(Z_{1:t}(x,x^-))= \{ x < x^- \}$ and $h(x)=-1\big]$ or $\big[\maj(Z_{1:t}(x,x^+))= \{ x > x^+ \}$ and $h(x)=1\big]$, \textbf{then} $\text{ANS} \assign \text{NO}$ and {\bfseries exit} loop. }

\ENDFOR

\STATE \textbf{If} $\maj\big(Z_{1:N}(x,x^-)\big) = \{ x > x^- \}$ and $\maj\big(Z_{1:N}(x,x^+)\big) = \{ x < x^+ \}$ \textbf{then} $\text{ANS} \assign \text{NO}$ and $S_{\text{in}} \assign S_{\text{in}} \cup \{x\}$.\label{step:filter-inside}

\STATE \textbf{If} $\text{ANS} = \text{YES}$, \textbf{then} $S_I \assign S_I \cup \{x\}$.

\ENDFOR

\STATE $\bar{S_{\text{in}}} \assign \complabel(S_{\text{in}}, \delta/3)$.

\STATE $S_I \assign S_I \cup \{x: (x, y) \in \bar{U} \cup \bar{S_{\text{in}}}\ \text{and}\ y \neq h(x) \}$.



{\bfseries return} $S_I$.

\end{algorithmic}	
\end{algorithm}

\subsection{Analysis of Phase 2} \label{subsec:phase_2}

Phase 2 aims to obtain instances obeying the distribution $\D_2$ as defined in Theorem~\ref{thm:boost}. To this end, we need to find a set $S_I$ of $\Theta(n_{\sqrt{\epsilon}})$ instances that $h_1$ will misclassify, and $\Theta(n_{\sqrt{\epsilon}})$ instances that $h_1$ will predict correctly. It then suffices to call the standard PAC learner $\A_\H$ to learn $h_2$. As we discussed in the preceding section, in order to guarantee the existence of $S_I$ with such size, we have to draw a set $S_2$ of $\Theta(n_{\epsilon})$ instances from $\D_X$. In order to identify them, we have to design a computationally efficient algorithm. We remind that, with $\Theta(n_{\epsilon})$ instances in $S_2$, we could not directly apply \complabel on $S_2$ to identify $S_I$ since it would lead to an undesirable overhead similar to that of the natural approach. We tackle this challenge by designing a novel algorithm termed \filter (Algorithm~\ref{alg:filter}). On the other hand, obtaining $S_C$ is relatively easy: since the error rate of $h_1$ is upper bounded by $\frac{\sqrt{\epsilon}}{2}$, it makes correct prediction on most of the instances. Therefore, it suffices to draw $\Theta(n_{\sqrt{\epsilon}})$ samples from $\D_X$ and call \complabel to obtain the correct label of all these data, which can be used to test the performance of $h_1$. By the Chernoff bound, it is guaranteed that we can find at least $(1 - \frac{\sqrt{\epsilon}}{2})n_{\sqrt{\epsilon}} \geq \frac12 n_{\sqrt{\epsilon}}$ such instances to form $S_C$. In the following, we elaborate on the design of the \filter algorithm, which is the key technical contribution of the paper.

From now on, we switch to the notation in Algorithm~\ref{alg:filter}. The principle of \filter is to utilize a modest amount of comparisons and a small amount of label queries to infer the correct label of a vast fraction of the instances of $S$ (i.e. the set $S_2$ in Algorithm~\ref{alg:boost}). To aid intuition, imagine that all the instances in $S$ are ordered by the true comparison function $Z^*(\cdot, \cdot)$ and then mapped into the interval $[0, 1]$ on the real line. There are two major stages in \filter: a sub-sampling stage to obtain ${U}$ and a filtering stage to obtain $S_I$.

In the sub-sampling stage, we uniformly sample a set $U \subset S$ and invoke the \complabel algorithm to obtain a correctly sorted and labeled set $\bar{U}$, enabling us to find two support instances $x^-$ and $x^+$. Roughly speaking, $x^-$ is the negative instance that is ``closest'' to positive instances in $U$, and $x^+$ is the positive instance that is ``closest'' to negative. 
One important consequence of this step is that the label of the instances outside the interval $[x^-, x^+]$ can be inferred provided that we have truthful comparison tags. On the other hand, we are uncertain about the label of those inside the interval. Since our objective is to collect a sufficient number of instances to form $S_I$, we need to constrain the size of the points residing in the interval. While it seems that a constant length (i.e. it contains a constant fraction of points in $S$) suffices, we will show that it needs to be as small as $\frac{\sqrt{\epsilon}}{4}$~--~this will be clarified when we explain the comparison complexity of the filtering step. Observe that such narrow interval can be guaranteed when the size of $U$ is as large as $\frac{4}{\sqrt{\epsilon}}\log\frac{16}{\delta}$; see Lemma~\ref{lem:size-inside-interval-restate} in Appendix~\ref{sec:app:phase2}.

The filtering stage is devoted to identifying good cases to test the performance of $h$. In particular, we hope to find instances whose true label can be inferred from a few noisy comparison tags, which can then be compared to the label predicted by $h$. In view of the availability of the support instances $x^-$ and $x^+$, it suffices to gather comparison tags for $(x, x^-)$ and $(x, x^+)$ for all $x \in S \backslash U$. When $x$ resides in the interval, we store it in $S_{\text{in}}$ and invoke \complabel on $S_{\text{in}}$ after all instances in $S\backslash U$ have been visited. This step is query-efficient since $U$ is constructed in such a way that $\abs{S_{\text{in}}} \leq O(\sqrt{\epsilon} \abs{S}) = O(n_{\sqrt{\epsilon}})$.



For those outside the interval,  we think of them as good cases to test $h$. This is because Proposition~\ref{prop:label-comp-complexity} guarantees that $U$ will be annotated correctly with high probability and hence, instances left to $x^-$ are likely to be negative (likewise instances right to $x^+$ are likely to be positive). In other words, the labels of $x^-$ and $x^+$ together with the comparison model we considered (see Section~\ref{sec:setup}) would help infer the label of the instances outside the interval; these {\em inferred labels} will then be contrasted to the predictions made by $h$. Indeed, Step~10 of \filter is exactly testing the matching between the inferred label and the predicted label from $h$. Such procedure would truthfully identify $S_I$ provided that the comparison to the support instances returned by the crowd has good quality.  In this regard, each instance $x$ will be assigned to multiple workers, and the underlying question is how many times are sufficient to justify the correctness of $h(x)$. A straightforward approach is to acquiring $N = O(\frac{1}{\beta^2} \log \frac{\abs{S}}{\delta})$ comparison tags per instance $x$ to guarantee that the majority votes $\maj(Z_{1:N}(x, x^-))$ and $\maj(Z_{1:N}(x, x^+))$ are correct for all $x \in S$ with high probability. This, however, leads to a comparison complexity bound of ${O}(\abs{S}\log\abs{S}) = {O}(n_{\epsilon}\log n_{\epsilon})$, which is comparable to the natural approach. In our algorithm, we explore the structure of the crowd and the error rate of the learned classifier to reduce it to ${O}(\abs{S}) = {O}(n_{{\epsilon}})$. In particular, since $\err_{\D_X}(h) = \Theta(\sqrt{\epsilon})$, the classifier essentially performs well on most of the data in $S$. Therefore, if there is one time stamp $t$ such that the inferred label matches the predicted label from $h$, then it implies that $h$ is correct on such instance $x$ with high probability. On the other hand, if after $N$ steps, the algorithm never found any group of workers whose majority vote combined with the label of $x^-$ and $x^+$ agrees with the predicted label from $h$, then this is strong evidence that $h$ misclassifies the current instance. Now we are in the position to spell out how to obtain the $\tilde{O}(n_{\sqrt{\epsilon}})$ comparison complexity. For those $x$ that $h$ classifies correctly, it follows that the algorithm will run only a few steps to find a group of workers whose comparison tags in allusion to the label of the support instances would agree with $h$. In fact, we show that for any such $x$, the number of comparison queries stays as a constant provided that $\beta \geq \Omega(1)$, say, 70\% of the workers are perfect when providing comparison tags. Since all of such $x$ consist of $1-\frac{\sqrt{\epsilon}}{2}$ fraction of $S$ (recall that $\err_{\D_X}(h) \leq \frac{\sqrt{\epsilon}}{2}$), the comparison complexity is $O(\abs{S})$. For those $x$ that $h$ misclassifies, the algorithm will need $N$ comparison queries; since they consist of $\frac{\sqrt{\epsilon}}{2}$ fraction, we need a total of $O(\sqrt{\epsilon} \abs{S} N)$ comparison queries. Lastly, for those inside the interval, the algorithm also runs $N$ steps; yet, as the sub-sampling stage guarantees that the fraction is $O(\sqrt{\epsilon})$, it ends up with $O(\sqrt{\epsilon} \abs{S} N)$ comparisons as well. 

We summarize the performance guarantee of \filter below.

\begin{lemma}\label{lem:SI}
Consider the \filter algorithm. Assume that  $U$ is correctly labeled and $\beta \geq c_0$ for some absolute constant $c_0 \in (0, 1/2]$. Consider any given instance $x \in S$ outside the interval $[x^-, x^+]$. If $h(x)=h^*(x)$, it will be added to $S_I$ with probability at most $\frac14\sqrt{\epsilon}$; if $h(x)\neq h^*(x)$, it goes to $S_I$ with probability at least $\frac{4c_0}{1+2c_0}$. For any instance $x \in S$ that falls into the interval $[x^-, x^+]$, it will be added to $S_I$ with probability at most $\frac14\sqrt{\epsilon}$.
\end{lemma}


\begin{proposition}\label{prop:filter-guarantee}
Consider the \filter algorithm with $\abs{S}= \Theta(n_{\epsilon,\delta})$. Assume $\beta \geq c_0$ for some absolute constant $c_0 \in (0, 1/2]$. Then, with probability at least $1-\delta$, the algorithm returns an instance set $S_I$ with size $\Theta(n_{\sqrt{\epsilon}, \delta})$. The label complexity is $O\big(\frac{1}{\alpha^2} \cdot \log n_{\sqrt{\epsilon},\delta} \cdot \log{\log n_{\sqrt{\epsilon},\delta}}\big)$, and the comparison complexity is $O\big( n_{\sqrt{\epsilon},\delta} \cdot \log^2 n_{\sqrt{\epsilon},\delta} + n_{\epsilon,\delta}\big)$.
\end{proposition}



These observations together lead to the following performance guarantee of Phase~2 in Algorithm~\ref{alg:boost}. Readers may refer to Appendix~\ref{sec:app:phase2} for a detailed proof.


\begin{proposition}\label{prop:err-h2}
Assume $\beta \geq c_0$ for some absolute constant $c_0 \in (0, 1/2]$. In Phase~2, with probability $1-\frac{\delta}{3}$, $\err_{\D_2}(h_2)\leq \frac{\sqrt{\epsilon}}{2}$. The label complexity is $O\big(\frac{1}{\alpha^2} \cdot \log n_{\sqrt{\epsilon},\delta} \cdot \log{\log n_{\sqrt{\epsilon},\delta}}\big)$, and the comparison complexity is $O\big(  n_{\sqrt{\epsilon},\delta} \cdot \log^2 n_{\sqrt{\epsilon},\delta} + n_{\epsilon,\delta} \big)$.
\end{proposition}

\subsection{Analysis of Phase 3}\label{subsec:phase_3}

Given the hypotheses $h_1$ and $h_2$, we can draw instances from the disagreement region via rejection sampling. 


\begin{proposition}\label{prop:err-h3}
In Phase~3, with probability $1-\frac{\delta}{3}$, $\err_{\D_3}(h_3) \leq \frac{\sqrt{\epsilon}}{2}$. In addition, the label complexity is $O\big(\frac{1}{\alpha^2}\cdot\log n_{\sqrt{\epsilon},\delta}\cdot\log{\log n_{\sqrt{\epsilon},\delta}}\big)$ and the comparison complexity is $O\big(\frac{1}{\beta^2}\cdot n_{\sqrt{\epsilon},\delta} \cdot\log^2 n_{\sqrt{\epsilon},\delta}\big)$.
\end{proposition}

Observe that now Theorem~\ref{thm:main} follows from Propositions~\ref{prop:err-h1}, \ref{prop:err-h2}, \ref{prop:err-h3}, and Theorem~\ref{thm:boost}; see Appendix~\ref{sec:app:summary} for the full proof.

%% file: conclusion.tex
\section{Conclusion}\label{sec:con}

We have shown that for any class of threshold functions that is efficiently PAC learnable under the standard setting, it is possible to efficiently learn it from the noisy crowd annotations, with an $o(1)$ labeling overhead and $O(1)$ comparison overhead. To this end, we have developed a set of new techniques, including a comparison-equipped labeling primitive and a label-efficient filtering process. It would be interesting to study agnostic crowdsourced PAC learning, or to study other query types \cite{balcan2012robust}.

%

%% file: appendix.tex
\section{From Perfect Workers to Reliable Workers}\label{sec:app:massart}

{\bfseries Crowd model with perfect workers.} 
All of the theoretical results in this paper are established based on the existence of perfect workers: $(\frac12 + \alpha)$ fraction of the pool of workers are perfect when providing labels, in the sense that they always label according to the perfect hypothesis $h^*$, while the remaining $(\frac12 - \alpha)$ may behave adversarially. Likewise, $(\frac12 +\beta)$ fraction always provide comparison tags according to $Z^*$ and the remaining $(\frac12 - \beta)$ may behave adversarially.

In this section, we show that it is easy to relax the requirement of perfectness to reliability. 

\begin{definition}[Reliable worker]
A reliable worker $t$ who provides labels is defined by a function $g_t: \X \rightarrow \Y$ satisfying the following condition: given any instance $x \sim D_X$,
\begin{equation}
\Pr\big( g_t(x) = h^*(x) \big) \geq 1 - \eta_t(x),
\end{equation}
where the function $\eta_t(x) \in [0, \eta_L]$ for some given parameter $\eta_L < 1/2$. Likewise, a reliable worker $t'$ who provides comparison tags is defined by a function $g_t': \X \times \X \rightarrow \{-1, 1\}$ satisfying the following condition: given any pair of instances $(x, x')$,
\begin{equation}
\Pr\big( g_t'(x, x) = Z^*(x, x') \big) \geq 1 - \eta_t'(x, x'),
\end{equation}
where the function $\eta_t'(x, x') \in [0, \eta_C]$ for some given parameter $\eta_C < 1/2$.
\end{definition}

Note that the function $\eta_t(x)$ (likewise for $\eta_t'(x, x')$), namely the probability that a worker $t$ will flip a given instance, may vary across workers and instances, and the only restriction is that the flipping probabilities are upper bounded by a parameter $\eta < 1/2$; thus this is a very flexible and strong noise model. In fact, such noise model is exactly the Massart noise~\cite{sloan1988types,massart2006risk} which has been broadly studied in the literature under the non-crowdsourcing setting and only until recently have polynomial-time algorithms been established under distributional assumptions \cite{zhang2020efficient,diakonikolas2020learning}; in addition, under the non-crowdsourcing setting, there is evidence that efficient learning with Massart noise without distributional assumptions is computationally hard~\cite{diakonikolas2020near}.

{\bfseries Crowd model with reliable workers.}
We consider the following crowd model: $\frac12 + \alpha'$ fraction of the workers are reliable while the remaining $\frac12 - \alpha'$ are adversarial for label queries. Likewise, $\frac12 + \beta'$ are reliable while the remaining $\frac12 - \beta'$ are adversarial for comparison queries.

We show that {\em all of our analysis based on the perfect version can be easily extended to the reliable version}.\footnote{The analysis of \citet{awasthi2017efficient} for $\alpha > 0$ holds for the reliable version of crowd model as well.} To simplify the discussion, we show how to do so for workers that provide labels; the case of comparison queries follows exactly the same reasoning. 

Indeed, all of our analysis uses the following fact from the perfect version of crowd: for any given instance $x$, by querying a randomly chosen worker, we are able to obtain the correct label with probability at least $\frac12 + \alpha$, namely
\begin{equation}\label{eq:con}
\Pr_{t \sim \PL}\big( h_t(x) = h^*(x) \big) \geq \frac12 + \alpha,
\end{equation}
where we recall that $\PL$ is the uniform distribution over the pool of workers that provide labels.

Now we can consider that the perfect workers are replaced with reliable workers, and it suffices to establish a condition similar to \eqref{eq:con}. 

Given any instance $x \sim D_X$, we have
\begin{align*}
\Pr_{t \sim \PL}\big( g_t(x) = h^*(x) \big) &\geq \Pr_{t \sim \P}\big( g_t(x) = h^*(x),\ t\ \text{ is\ reliable} \big)\\
&= \Pr_{t \sim \P}\big( g_t(x) = h^*(x) \mid t\ \text{ is\ reliable} \big) \cdot \Pr_{t \sim \P}\big( t\ \text{ is\ reliable} \big) \\
&\geq \big(1 - \eta_t(x) \big) \cdot \Big(\frac12 + \alpha' \Big)\\
&\geq (1-\eta) \cdot \Big(\frac12 + \alpha' \Big)\\
&= \frac12 + \Big[ (1-\eta) \cdot \Big(\frac12 + \alpha' \Big) - \frac12 \Big].
\end{align*}
Therefore, under the reliable version of the crowd model, all of our analysis holds but with a new parameter $\alpha = (1-\eta) \cdot \big(\frac12 + \alpha' \big) - \frac12$, where the parameters $\eta \in [0, 1/2)$ and $\alpha' \in (0, 1/2]$ are such that $(1-\eta) \cdot \big(\frac12 + \alpha' \big) - \frac12 \in (0, 1/2]$.

Similarly, the assumption that $\frac12 + \beta$ fraction of workers are perfect when providing comparison tags can be relaxed to that $\frac12 + \beta'$ fraction are reliable, and all of our analysis holds but with a new parameter $\beta = (1-\eta) \cdot \big(\frac12 + \beta' \big) - \frac12$, where the parameters $\eta \in [0, 1/2)$ and $\beta' \in (0, 1/2]$ are such that $(1-\eta) \cdot \big(\frac12 + \beta' \big) - \frac12 \in (0, 1/2]$.

\section{Omitted Proof from Section~\ref{subsec:natural-approach}}\label{sec:app:comp-and-label}

\begin{proposition}[Restatement of Proposition~\ref{prop:label-comp-complexity}]\label{prop:label-comp-complexity-restate}
Consider the \complabel algorithm, i.e. Algorithm~\ref{alg:label}. If $\abs{S} \geq  (\frac{3}{\delta})^{1/1000}$, then with probability at least $1-\delta$, it correctly sorts and labels all the instances in $S$. The label complexity is $O\big(\frac{1}{\alpha^2}\cdot \log\abs{S}\cdot\log{\log\abs{S}} \big)$, and the comparison complexity is given by $O\big(\frac{1}{\beta^2}\cdot\abs{S}\cdot\log^2\abs{S}\big)$.
\end{proposition}

\begin{proof}
Recall that given any pair of instances $(x,x')\in\X\times\X$, if we randomly draw a worker $t\sim\PC$, with probability at least $\frac{1}{2}+\beta$, the comparison tag is correct. 

Let $k_1$ be the number of workers we query for each pair of instances in Algorithm~\ref{alg:label}. The probability that the majority of the $k_1$ tags is incorrect on a given pair $(x,x')$ is

\[
\Pr\Big(Z^*(x,x') \cdot \sum_{j=1}^{k_1} Z_t(x,x') \leq 0\Big),
\]
where $Z_t(x, x')$ is the annotation from the $j$-th worker.

Without loss of generality, we assume the ground truth $Z^*(x,x') = -1$. It then follows that $\E_{t \sim \P_C}\big[Z_t(x,x')\big] = -2\beta < 0$. By Hoeffding's inequality, we have
\begin{equation}\label{eq:hoeff_1}
\Pr\bigg(\sum_{t=1}^{k_1} Z_t(x,x') \geq
k_1\cdot\E_t\big[Z_t(x,x')\big] + t \bigg) \leq
e^{-\frac{2t^2}{k_1\cdot (b-a)^2}},
\end{equation}
where $a=-1$ and $b=1$ are the corresponding lower and upper bounds of each tag $Z_t(x,x')$. Let $t$ be such that $k_1\cdot\E_t\big[Z_t(x,x')\big] + t = 0$. Then
\begin{equation}
\Pr\bigg(\sum_{t=1}^{k_1} Z_t(x,x') \geq
0 \bigg) \leq
e^{-{2k_1\beta^2}}.
\end{equation}

Let $q_S$ be the total number of comparisons made by algorithm \quicksort, and denote by $(x_l, x_l')$ the pair being compared in the $l$-th iteration. We apply the union bound and obtain that
\begin{equation}\label{eq:union_1}
\Pr\bigg(\bigcup_{l=1}^{q_S} \bigg[ Z^*(x_l,x_l')\sum_{t=1}^{k_1} Z_t(x_l,x_l') \leq 0 \bigg]\bigg)\leq
\sum_{l=1}^{q_S} e^{-2k_1\beta^2} \leq
q_S \cdot e^{-2k_1\beta^2}.
\end{equation}
By setting
\begin{equation*}
k_1 =  \frac{1}{2\beta^2}\cdot \log\frac{3q_S}{\delta},
\end{equation*}
we have that with probability at least $\frac{\delta}{3}$, the set $S$ is correctly sorted.

Now we upper bound the quantity $q_S$. Recall that by Lemma~\ref{lem:quicksort}, with probability at least $1-\frac{1}{\abs{S}^c}$, we have $q_S \leq (c+2) \abs{S} \log_{8/7}\abs{S}$. By our requirement on the size of $S$, $c=1000$. Thus, with probability at least $1-\frac{\delta}{3}$, $q_S \leq 1002 \abs{S} \log_{8/7}\abs{S}$. Thus, it suffices to set
\begin{equation}
k_1 = \frac{1}{2\beta^2} \cdot \log \frac{3006\abs{S} \log_{8/7}\abs{S}}{\delta}
\end{equation}
to ensure that with probability at least $1-\frac{2}{3}\delta$, the set $S$ is correctly sorted. It is not hard to see that the number of comparisons is upper bounded by
\begin{equation}
k_1 \cdot q_S \leq \frac{1}{2\beta^2} \cdot \log \frac{3006\abs{S} \log_{8/7}\abs{S}}{\delta} \cdot 1002 \abs{S} \cdot \log_{8/7}\abs{S}  \leq O\Big( \frac{1}{\beta^2} \abs{S} \cdot \log^2\abs{S}  \Big).
\end{equation}

Likewise, for any given instance $x\in\X$, let $g_i(x)$ be the label given by a randomly selected worker $i\sim\PL$. Recall that $g_i(x) = h^*(x)$ with probability at least $\frac{1}{2}+\alpha$. Similar to the previous analysis, by Hoeffding's inequality and the union bound (over the labeling of the $\log\abs{S}$ instances), we have that
\begin{equation*}
\Pr\bigg(\bigcup_{l=1}^{\log\abs{S}} \bigg[h^*(x_l) \sum_{i=1}^{k_2}g_i(x_l)\leq 0 \bigg] \bigg) 
\leq \sum_{l=1}^{\log \abs{S}} \Pr \bigg( h^*(x_l)\sum_{i=1}^{k_2}g_i(x_l)\leq 0 \bigg) 
\leq \log \abs{S} \cdot e^{-{2k_2\alpha^2}{}} = \frac{\delta}{3},
\end{equation*}
provided that we set
\begin{equation}
k_2 = \frac{1}{2\alpha^2} \cdot \log\frac{3\log\abs{S}}{\delta}.
\end{equation}
The total number of calls to the labeling oracle equals
\begin{equation}
k_2 \cdot \log\abs{S} =  \frac{1}{2\alpha^2} \cdot \log\frac{3\log\abs{S}}{\delta} \cdot \log\abs{S} = O\Big( \frac{1}{\alpha^2} \cdot \log\abs{S} \cdot \log{\log\abs{S}} \Big).
\end{equation}

By union bound, we have that with probability at least $1-\delta$, $S$ is correctly sorted and labeled, which proves the desired result.
\end{proof}


\begin{theorem}[Restatement of Theorem~\ref{thm:natural-approach}]\label{thm:natural-approach-restate}
With probability at least $1-\delta$, the natural approach runs in time $\poly(d,\frac{1}{\epsilon})$ and returns a classifier $h$ with error rate $\err_{\D_X}(h)\leq\epsilon$. The label complexity is $O\big(\frac{1}{\alpha^2} \cdot \log n_{\epsilon,\delta} \cdot \log {\log n_{\epsilon,\delta}}\big)$ and the comparison complexity is $O\big( \frac{1}{\beta^2} n_{\epsilon,\delta} \cdot \log^2 n_{\epsilon,\delta} \big)$. Therefore, the labeling overhead $\labeloverhead= \frac{1}{\alpha^2} \cdot \tilde{O}\big( \frac{\log(d/\epsilon)}{d/\epsilon} \big)$ and the comparison overhead $\compareoverhead= O_{\delta}\big( \frac{1}{\beta^2} \cdot \log^2 n_{\epsilon,\delta} \big)$.
\end{theorem}
\begin{proof}
We note that by \eqref{eq:m_eps}, $\abs{S} = n_{\epsilon,\delta} \geq (\frac{3}{\delta})^{1/1000}$. Thus, we can apply Proposition~\ref{prop:label-comp-complexity-restate} to show that with probability $1-\delta$, the set $S$ is correctly labeled by Algorithm~\ref{alg:label}. This in allusion to Assumption~\ref{as:pac} implies the PAC guarantee of the natural approach.

Now we give the labeling and comparison overhead. To this end, we first note that the label and comparison complexity inherits from Proposition~\ref{prop:label-comp-complexity-restate}. Thus, we have the label complexity and comparison complexity as follows:
\begin{align}
m_L &= O\Big(\frac{1}{\alpha^2} \cdot \log n_{\epsilon,\delta} \cdot \log {\log n_{\epsilon,\delta}} \Big)= \frac{1}{\alpha^2} \cdot \tilde{O}\Big( \log\frac{d+ \frac{1}{\delta}}{\epsilon} \Big),\\
m_C &= O\Big( \frac{1}{\beta^2} n_{\epsilon,\delta} \cdot \log^2 n_{\epsilon,\delta} \Big).
\end{align}
In the above expression, the $\tilde{O}(\cdot)$ notation hides poly-logarithmic factors of the argument.

Therefore, by $m_{\epsilon, \delta} \geq K \cdot \frac{d + \log(1/\delta)}{\epsilon}$, we have
\begin{equation}
\labeloverhead = \frac{m_L}{m_{\epsilon,\delta}} \leq \frac{1}{\alpha^2} \cdot \tilde{O}\Big( \frac{\epsilon}{d + \log(1/\delta)} \cdot \log\frac{d+1/\delta}{\epsilon} \Big) = \frac{1}{\alpha^2} \cdot \tilde{O}\Big( \frac{\log(d/\epsilon)}{d/\epsilon} \Big),
\end{equation}
which goes to zero as $\epsilon$ goes to $0$.

For the comparison overhead, we have
\begin{equation}
\compareoverhead = O\Big( \frac{1}{\beta^2} \cdot \frac{n_{\epsilon,\delta} \log^2 n_{\epsilon,\delta}}{m_{\epsilon,\delta}} \Big) = O_{\delta}\Big( \frac{1}{\beta^2} \cdot \log^2 n_{\epsilon,\delta} \Big).
\end{equation}

To see that the algorithm runs in polynomial time, first, we recall that Algorithm~\ref{alg:label} runs in polynomial time. In addition, the standard PAC learner $\A_\H$ is assumed to run in polynomial time. The proof is complete.
\end{proof}

\section{Analysis of Phase 1}\label{sec:app:phase1}

\begin{proposition}[Restatement of Prop.~\ref{prop:err-h1}]\label{prop:err-h1-restate}
In Phase~1, with probability $1-\frac{\delta}{3}$, the classifier $h_1$ is such that $\err_{\D_X}(h_1) \in \big[\frac16\sqrt{\epsilon}, \frac12\sqrt{\epsilon}\big]$. In addition, the label complexity is $O\big(\frac{1}{\alpha^2}\cdot\log n_{\sqrt{\epsilon},\delta}\cdot\log{\log n_{\sqrt{\epsilon},\delta}}\big)$
and the comparison complexity is $O\big(\frac{1}{\beta^2}\cdot n_{\sqrt{\epsilon},\delta} \cdot\log^2 n_{\sqrt{\epsilon},\delta}\big)$.
\end{proposition}
\begin{proof}
Recall that $\abs{S_1}= n_{\sqrt{\epsilon}/2,{\delta}/{6}}$. 
By Proposition~\ref{prop:label-comp-complexity-restate}, all the instances in $\bar{S_1}$ are correctly labeled with probability $1-\frac{\delta}{6}$. Thus, conditioned on this event, with probability at least $1-\frac{\delta}{6}$, $\A_\H(\bar{S_1}, \frac{\sqrt{\epsilon}}{2}, \frac{\delta}{6})$ returns a classifier $h_1'$ with $\err_{\D_X}(h_1') \leq \frac12\sqrt{\epsilon}$. By union bound (over the labeling of $S_1$ and the learner $\A_\H$), we know that with probability at least $1-\frac{\delta}{3}$, $\err_{\D_X}(h_1') \leq \frac12\sqrt{\epsilon}$. Observe that the classifier $h_1'$ corresponds to the parameters $\eta = \frac12\sqrt{\epsilon}$ and $c = \frac12$ in Algorithm~\ref{alg:anti}. Hence, by Lemma~\ref{lem:anti}, the obtained classifier $h_1$ is such that $\err_{\D_X}(h_1) \in \big[ \frac16\sqrt{\epsilon}, \frac12\sqrt{\epsilon}\big]$.

Note that we only query workers when training $h_1'$. Hence, by Proposition~\ref{prop:label-comp-complexity-restate}, the label complexity is $O\big(\frac{1}{\alpha^2}\cdot\log n_{\sqrt{\epsilon},\delta}\cdot\log\frac{\log n_{\sqrt{\epsilon},\delta}}{\delta}\big)$ and the comparison complexity is $O\big(\frac{1}{\beta^2}\cdot n_{\sqrt{\epsilon},\delta} \cdot\log n_{\sqrt{\epsilon},\delta}\cdot\log\frac{n_{\sqrt\epsilon,\delta}}{\delta}\big)$.
\end{proof}


\begin{lemma}\label{lem:anti}
Consider Algorithm~\ref{alg:anti}. The returned classifier $h: \X \rightarrow \Y$ satisfies $\err_{D_X}(h) \in [c_1\eta, c_2\eta]$, where $c_1 = \min\big\{\frac12, \frac{1-c}{2-c}\big\}$ and $c_2 = \max\big\{1, c+\frac12 \big\}$.
\end{lemma}
\begin{proof}
Denote $\xi := \err_{D_X}(h')$. Observe that
\begin{align*}
&\quad\Pr_{x \sim D_X}(h(x) \neq h^*(x)) \\
&= \Pr_{x \sim D_X}(h(x) \neq h^*(x) \mid \text{HEADS}) \cdot \Pr(\text{HEADS}) + \Pr_{x \sim D_X}(h(x) \neq h^*(x) \mid \text{TAILS}) \cdot \Pr(\text{TAILS})\\
&= \Pr_{x \sim D_X}(h'(x) \neq h^*(x) \mid \text{HEADS}) \cdot \Pr(\text{HEADS}) + \Pr_{x \sim D_X}(-h'(x) \neq h^*(x) \mid \text{TAILS}) \cdot \Pr(\text{TAILS})\\
&= \xi \cdot \lambda + (1 - \xi) (1 - \lambda).
\end{align*}

First, we consider that $\xi < \frac12\eta$.
Note that since $\xi \cdot \lambda \geq 0$ and $\xi < \frac12\eta$, we have
\begin{equation}
\xi \cdot \lambda + (1 - \xi) (1 - \lambda) \geq 0 + (1 - \frac12\eta)(1-\lambda)
= \frac12 \eta,
\end{equation}
where the last step follows from our choice of $\lambda$. On the other hand, by using the fact that $0 \leq \xi < \frac12\eta$, we have
\begin{align}
\xi \cdot \lambda + (1 - \xi) (1 - \lambda) \leq \frac12\eta \cdot \lambda + (1 - \lambda) = \frac{1}{2}\eta\big( 1 - \frac{ \frac12\eta }{1-\frac12\eta} \big) +  \frac{\frac12\eta}{1-\frac12\eta} = \eta.
\end{align}
Therefore, we prove that the error rate of $h$ falls into the range of $\big[\frac12\eta, \eta\big]$ when $\xi < \frac12\eta$.

Now, we consider that in reality, $\xi \in [\frac12\eta, \eta$]. In this case, we have
\begin{equation}
\xi \cdot \lambda + (1 - \xi) (1 - \lambda) \geq 0 + (1 - \eta)(1-\lambda) = \frac{1-\eta}{2-\eta} \cdot \eta \geq \frac{1-c}{2-c} \cdot \eta.
\end{equation}
On the other hand, we also have
\begin{equation}
\xi \cdot \lambda + (1 - \xi) (1 - \lambda) \leq \eta \lambda + (1-\frac12\eta)(1-\lambda) = \eta \lambda + \frac12\eta \leq (c + \frac12) \eta,
\end{equation}
where in the equality we use the setting of $\lambda$. Therefore, we prove that when $\frac12\eta \leq \xi \leq \eta$, we still have $\err_{D}(h) \in \big[ \frac{1-c}{2-c}\eta, (c+\frac12)\eta \big]$.

The result follows by taking the union of the intervals obtained for the two cases of $\xi$.
\end{proof}


\section{Analysis of Phase 2}\label{sec:app:phase2}

We will first analyze the performance of the \textsc{Filter} algorithm, i.e. Algorithm~\ref{alg:filter}. Then, we give a full analysis for Phase 2 of Algorithm~\ref{alg:boost}.

\begin{figure}[h]
\centering
\includegraphics[width=0.7\columnwidth]{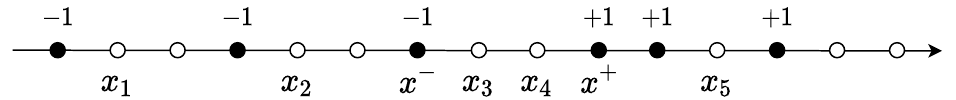}
\caption{{\bfseries Illustration of \filter.} Each circle represents an instance in $S$, where we use solid circles to indicate the instances in $U \subset S$. We call the oracles to sort and label all the instances in $U$ correctly. We use $x^-$ to denote the rightmost instance in $U$ that is labeled as $-1$, and use $x^+$ to denote the leftmost instance in $U$ that is labeled as $+1$. We will think of $[x^-, x^+]$ as an interval, and instances outside the interval (e.g. $x_1$, $x_5$) are potentially informative, since we can combine the label of $x^-$ or $x^+$ and the comparison tag to infer their labels (e.g. $x_5$ is likely positive). Thus, these points will be used to test the performance of $h$.
}
\label{fig:alg3}
\end{figure}

\subsection{Analysis of \textsc{Filter}}\label{sec:app:filter}

We will frequently discuss instances outside the interval $[x^-, x^+]$, which can be formally defined as follows: for the input set $S$ in \textsc{Filter}, the perfect comparison function $Z^*(\cdot, \cdot)$ gives a full ranking of all the instances. We then  map these points to the interval $[0, 1]$ of the real line by their order. An instance $x$ is called outside the interval $[x^-, x^+]$ if its coordinate in the real line is less than that of $x^-$ or greater than that of $x^+$. Note that this is for illustration purpose; in our algorithm we have no access to $Z^*(\cdot, \cdot)$.

In the subsequent analysis, we use the notation $(S, h, \delta)$ to denote the inputs to \filter, which corresponds to the arguments $(S_2, h_1, \delta/12)$ of \filter when invoked in Algorithm~\ref{alg:boost}. We recall that $\abs{S} = \Theta(n_{\epsilon, \delta})$ and $\err_{\D_X}(h) = \Theta(\sqrt{\epsilon})$.


\subsubsection{Analysis of sub-sampling}

We first show that with high probability, $U$ is correctly sorted and labeled. Then we will condition on this event and analyze the remaining steps of \textsc{Filter}.

\begin{lemma} \label{lem:subsample}
Consider \filter (Algorithm ~\ref{alg:filter}) with input $(S, h, \delta)$. With probability $1-\delta/8$, Algorithm~\ref{alg:filter} correctly sorts and labels the subset $U$. The label complexity of the sub-sampling step is 
$O\big(\frac{1}{\alpha^2}\cdot \log \abs{U} \cdot\log{\log \abs{U}} \big)$, and the comparison complexity of the sub-sampling step is $O\big(\frac{1}{\beta^2}\cdot \abs{U} \cdot\log^2 \abs{U}\big)$.
\end{lemma}
\begin{proof}
Note that $\abs{U} = \frac{4}{\sqrt{\epsilon}} \log\frac{16}{\delta} + (\frac{24}{\delta})^{1/1000} \geq (\frac{24}{\delta})^{1/1000}$. Hence, the results follow from Proposition~\ref{prop:label-comp-complexity-restate}.
\end{proof}

\begin{lemma}\label{lem:size-inside-interval-restate}
Consider Algorithm~\ref{alg:filter}. Suppose that the subset $U$ is correctly labeled. With probability at least $1- \delta/8$, the interval $[x^-, x^+]$ contains less than $\eta$ fraction of the instances in $S$ provided that $\abs{U} \geq \frac{1}{\eta} \log\frac{16}{\delta}$.
\end{lemma}
\begin{proof}
The algorithm uniformly samples a subset $U\subseteq S$ with $b$ instances. Without loss of generality, let $x_1<x_2<\dots<x_b$ be the instances in $U$.

Denote by $E_i$ the event that interval $[x_i,x_{i+1}]$ contains more than a $\eta$ fraction of the instances in $S$. Let $\rho$ be the distance from $x_i$ to the leftmost instance $x_0$ (note that all the points are  evenly mapped to the interval $[0, 1]$ and hence $\rho$ is exactly the coordinate of $x_i$ after mapping). Let $L$ be the set of instances that are left to $x_i$, and $R$ be those right to $x_{i+1}$. Then, $E_i$ happens only if the distance of all cross-group points (i.e. for all $x \in L$ and $x' \in R$ ) is at least $\eta$. 
In the following, we only consider the case that $\rho \in [0, 1-\eta]$, because otherwise the fraction of the data contained in $[x_i,x_{i+1}]$ must be less than $\eta$. Therefore, for any $ 1 \leq i \leq b-1$, we have
\begin{equation*}
\Pr\big(E_i\big) \leq \int_{0}^{1-\eta} \rho^{i-1} \cdot \big(1 - \eta - \rho\big)^{b-i} d\rho \leq \frac{2(1-\eta)}{b+1} \cdot \big( 1-\eta \big)^{b} \leq \frac{2}{b+1} e^{-b\eta},
\end{equation*}
where the last step follows from the inequality $1 - t \leq e^{-t}$ for all $t \geq 0$ and the fact that $1 - \eta \leq 1$.

By union bound over the intervals, we have
\begin{equation*}
\Pr\Big(\bigcup_{i=1}^{b-1}E_i\Big) \leq \frac{2(b-1)}{b+1} \cdot e^{-b\eta} \leq 2 \cdot e^{-b\eta}.
\end{equation*}
Recall that we set the parameter $b \geq \frac{1}{\eta} \log\frac{16}{\delta}$, which implies that the above probability is upper bounded by $\delta/8$. Namely, the probability that the interval $[x^-,x^+]$ contains less than $\eta$ fraction of the instances of $S$ is at least $1-\delta/8$.
\end{proof}

\subsubsection{Analysis of the structure of $S_I$}\label{sec:app:SI}

\begin{lemma}[Restatement of Lemma~\ref{lem:SI}]\label{lem:SI-restate}
Consider Algorithm~\ref{alg:filter}. Assume that the subset $U$ is correctly labeled and $\beta \geq c_0$ for some absolute constant $c_0 \in (0, 1/2]$. Consider any given instance $x \in S$ outside the interval $[x^-, x^+]$. If $h(x)=h^*(x)$, it will be added to $S_I$ with probability at most $\frac14\sqrt{\epsilon}$; if $h(x)\neq h^*(x)$, it goes to $S_I$ with probability at least $\frac{4c_0}{1+2c_0}$. For any instance $x \in S$ that falls into the interval $[x^-, x^+]$, it will be added to $S_I$ with probability at most $\frac14\sqrt{\epsilon}$.
\end{lemma}
\begin{proof}
Consider the following two events that were involved in the Filtering step of the algorithm \textsc{Fliter}: $(A_t)$ $\maj(Z_{1:t}(x, x^-)) = \{x < x^-\}$ and $h(x) = -1$; $(B_t)$ $\maj(Z_{1:t}(x, x^+)) = \{x > x^+\}$ and $h(x) = 1$. 

\vspace{0.1in}
\noindent{\bfseries Case 1.} The instance $x$ is outside the interval and $h(x) = h^*(x)$. By assumption, both $x^-$ and $x^+$ are labeled correctly. On the other hand, the probability that the comparison tag from a random worker $t\sim\PC$ is correct is at least $\frac12+\beta$. Thus, by the Chernoff bound, the majority vote of $N=\frac{1}{\beta^2}\log\frac{16}{\epsilon}$ comparison tags is correct with probability at least $1- \frac14\sqrt{\epsilon}$, i.e. $\maj(Z_{1:N}(x, x^-)) = Z^*(x, x^-)$ and $\maj(Z_{1:N}(x, x^+)) = Z^*(x, x^+)$. Consider that in reality, $Z^*(x, x^-) = \{x < x^-\}$. This immediately implies $\maj(Z_{1:N}(x, x^-)) = \{x < x^-\}$. On the other hand, this in allusion to the comparison model in Section~\ref{sec:setup} also implies $h^*(x) = -1$. Since we are considering $x$ such that $h(x) = h^*(x)$, we have $h(x) = -1$. Therefore, the event $A_N$ occurs. Similarly, we can show that if in reality, $Z^*(x, x^+) = \{x > x^+\}$, then the event $B_N$ occurs. Together, we have that either $A_t$ or $B_t$ occurs when $t = N$ if not earlier, and hence we will set ANS to NO and such instance will not be added to $S_I$.

\vspace{0.1in}
\noindent{\bfseries Case 2.} The instance $x$ is outside the interval and $h(x)\neq h^*(x)$. Let $E_t = A_t \cup B_t$. We aim to show that with constant probability, $E_t^c$, the complement of $E_t$, occurs for all $t \leq N$, hence $x$ will be added to $S_I$. To this end, it suffices to show that there exists at least a time stamp $t \leq N$, such that $E_t$ happens with probability at most $O(1)$, because $\Pr(\cap E_t^c) \geq 1 - \Pr(\cup E_t)$. Observe we can rewrite $A_t$ and $B_t$ in conjunction with the condition $h(x) \neq h^*(x)$ as follows: $(A_t)$ $\maj(Z_{1:t}(x, x^-)) = \{x < x^-\}$ and $h^*(x) = 1$; $(B_t)$ $\maj(Z_{1:t}(x, x^+)) = \{x > x^+\}$ and $h^*(x) = -1$. Recall that when either $A_t$ or $B_t$ occurs, it must be the case that the majority vote of the $t$ workers is incorrect, where we defined the incorrectness of a comparison tag in Section~\ref{sec:setup}. It hence remains to upper bound the probability of such event. This can essentially be formed as a biased random walk, also known as the probability of ruin in gambling~\cite{feller2008introduction}: we are given a random walk that takes a step to the right with probability $\frac12+\beta$ (corresponding to a draw of a perfect worker) and takes a step to the left with probability $\frac12-\beta$, and the question is how likely the walk will ever cross the origin to the left while taking $N$ steps. By Lemma~\ref{lem:ruin}, we know that such probability is given by
\begin{equation}
\frac{1 - \big(\frac{1/2+\beta}{1/2-\beta}\big)^N }{1- \big( \frac{1/2 + \beta}{1/2 - \beta} \big)^{N+1} } \leq \frac{\frac12 - \beta}{\frac12 + \beta} \leq \frac{1 - 2 c_0}{1 + 2 c_0},
\end{equation}
provided that $\beta \geq c_0$ for some absolute constant $c_0 > 0$. Therefore, with probability at least $\frac{4c_0}{1+2c_0}$, $x$ will be added to $S_I$.

\vspace{0.1in}
\noindent{\bfseries Case 3.} The instance $x$ falls into the interval. Using same argument as in Case 1, we know that the majority vote of $N$ workers agrees with $Z^*$ with probability at least $1-\frac14\sqrt{\epsilon}$. Therefore, such instance must be detected after the $N$-th iteration.
\end{proof}

We now have the following corollary regarding the size of $S_I$.
\begin{corollary}\label{coro:SI-size}
Consider Algorithm~\ref{alg:filter} with $\abs{S} = \Theta(n_{\epsilon, \delta})$. Assume that the subset $U$ is correctly labeled and $\beta \geq c_0$ for some absolute constant $c_0 \in (0, 1/2]$. With probability at least $1-\delta/4$, $\abs{S_I} = \Theta(\sqrt{\epsilon} \abs{S})$.
\end{corollary}
\begin{proof}
For any $x \in S$, let $r_x$ be a random variable that takes value $0$ if $x$ was not added to $S_I$ and takes value $1$ otherwise. 

Consider any instance $x \in S$ that is outside the interval $[x^-, x^+]$. By Lemma~\ref{lem:SI-restate}, we have
\begin{align*}
\Pr(r_x = 1)=&\ \Pr(r_x = 1 \mid h(x) = h^*(x) ) \cdot \Pr(h(x) = h^*(x)) \\
&\ + \Pr(r_x = 1 \mid h(x) \neq h^*(x) ) \cdot \Pr( h(x) \neq h^*(x))\\
\leq&\  \frac14\sqrt{\epsilon} \cdot 1 + 1 \cdot \frac12\sqrt{\epsilon} \leq  \sqrt{\epsilon}.
\end{align*}
On the other hand, it is not hard to see that
\begin{equation*}
\Pr(r_x = 1) \geq 0 + \frac{4c_0}{1+2c_0} \cdot \frac16\sqrt{\epsilon} = \Theta(\sqrt{\epsilon}).
\end{equation*}
Therefore, by the Chernoff bound, with probability $1 - e^{\Theta(-\sqrt{\epsilon} \abs{S})}$, we have $\abs{S_I} = \Theta(\sqrt{\epsilon}(\abs{S} - \abs{S_{\text{in}}}) + M)$, where $M$ denotes the number of instances we added from $U \cup S_{\text{in}}$. Now by Lemma~\ref{lem:size-inside-interval-restate} and our setting with $\abs{U} \geq \frac{4}{\sqrt{\epsilon}}\log\frac{16}{\delta}$, we have $0 \leq \abs{S_{\text{in}}} \leq \frac14\sqrt{\epsilon}\abs{S} \leq \frac14\abs{S}$ with probability at least $1-\frac{\delta}{8}$. In addition, $0 \leq M \leq \abs{U} + \abs{S_{\text{in}}} = \Theta(\sqrt{\epsilon}\abs{S})$. These together with $\abs{S} \geq \Omega(\frac{1}{\epsilon} \log\frac{8}{\delta})$ implies that with probability $1-\delta/4$, $\abs{S_I} = \Theta(\sqrt{\epsilon}\abs{S})$.
\end{proof}

\subsubsection{Performance guarantee of \textsc{Filter}}

A naive worst-case analysis would give query complexity bound of $O(\abs{S} N)$. In the following, we show an improved result. The proof follows closely from \citet{awasthi2017efficient}, where the new ingredient is that we need to examine the instances within and outside the interval respectively. This is where Lemma~\ref{lem:size-inside-interval-restate} plays a role in controlling the overall query complexity.

\begin{proposition}[Restatement of Prop.~\ref{prop:filter-guarantee}]\label{prop:filter-guarantee-restate}
Consider the \filter algorithm with $\abs{S}= \Theta(n_{\epsilon,\delta})$. Assume that $\beta \geq c_0$ for some absolute constant $c_0 \in (0, 1/2]$. Then, with probability at least $1-\delta$, the algorithm returns an instance set $S_I$ with size $\Theta(n_{\sqrt{\epsilon}, \delta})$. The label complexity is $O\big(\frac{1}{\alpha^2} \cdot \log n_{\sqrt{\epsilon},\delta} \cdot \log{\log n_{\sqrt{\epsilon},\delta}}\big)$, and the comparison complexity is $O\big(n_{\sqrt{\epsilon},\delta} \cdot \log^2 n_{\sqrt{\epsilon},\delta} + n_{\epsilon, \delta} \big)$.
\end{proposition}
\begin{proof}
First, by Lemma~\ref{lem:subsample} and Lemma~\ref{lem:size-inside-interval-restate} (and the setting of $\abs{U}$), with probability $1-\delta/3$, the sub-sampling step identifies $x^-$ and $x^+$ such that: 1) $x^-$ is negative and $x^+$ is positive; and 2) the interval $[x^-, x^+]$ contains less than $\frac14\sqrt{\epsilon}\abs{S}$ instances. In addition, since $\abs{U} \leq O(n_{\sqrt{\epsilon},\delta})$, the label complexity and comparison complexity of sub-sampling are $O\big(\frac{1}{\alpha^2} \cdot \log n_{\sqrt{\epsilon},\delta} \cdot \log{\log n_{\sqrt{\epsilon},\delta}}\big)$ and $O\big( \frac{1}{\beta^2} n_{\sqrt{\epsilon},\delta} \cdot \log^2 n_{\sqrt{\epsilon},\delta} \big)$ respectively.

We condition on these happening and consider comparison complexity for three (overlapping) cases for $x \in S$: it falls into the interval $[x^-, x^+]$, it is such that $h(x) \neq h^*(x)$, it is such that $h(x) = h^*(x)$.

\vspace{0.1in}
\noindent{\bfseries Case 1.} 
For the instances inside the interval, the total number of comparison queries on these points is upper bounded by $\frac14\sqrt{\epsilon} \abs{S}N$.

\vspace{0.1in}
\noindent{\bfseries Case 2.}
Since the error rate of $h$ is $\Theta(\sqrt{\epsilon})$, we know that with probability at least $1 - e^{-\sqrt{\epsilon}\abs{S}} \geq 1 - \delta/8$, the number of instances in $S$ on which $h$ disagrees with $h^*$ is $\Theta(\sqrt{\epsilon}\abs{S})$. Hence, the total number of comparison queries is $\Theta(\sqrt{\epsilon} \abs{S}N)$.

\vspace{0.1in}
\noindent{\bfseries Case 3.}
Lastly, we consider $x\in S$ such that $h(x)=h^*(x)$. Let $N_i$ be the expected number of queries we need until having $i$ more correct comparison tags than the incorrect ones. Then,
\begin{equation}\label{eq:N_1<N_2}
N_1\leq \big(\frac12+\beta \big)\cdot 1+ \big(\frac12-\beta \big)(N_2+1).
\end{equation}
This is because with probability at least $\frac12+\beta$, we obtain a correct comparison tag and stop, and with probability at most $\frac12 - \beta$, we get an incorrect tag and in the future, we must query the workers until we see $2$ more correct tags than incorrect ones. On the other hand, it is not hard to see that $N_2 = 2 N_1$, since in order to reach the status of $N_2$, we just repeat the process of getting $N_1$ twice. This combined with \eqref{eq:N_1<N_2} implies that
\[
N_1 \leq \frac{1}{2\beta} \leq \frac{1}{2c_0}
\]
as long as $\beta \geq c_0$. Now using the Bernstein's inequality (see e.g. Appendix~D of \citet{awasthi2017efficient}), we know that with probability at least $1 - e^{-\abs{S}} \geq 1 - \delta/8$, the total number of comparison queries on all $x$ with $h(x) = h^*(x)$ is $O(\abs{S})$.

Combining all three cases and the fact that $\abs{S} = \Theta(n_{\epsilon, \delta})$ and $N = \frac{1}{\beta^2}\log\frac{1}{\epsilon}$, we have that with probability $1- \delta/4$, the comparison complexity of the ``for $x \in S\backslash U$'' loop is
\begin{equation}
\frac14\sqrt{\epsilon} \abs{S} N + \Theta(\sqrt{\epsilon} \abs{S} N) + O(\abs{S}) \leq O(\abs{S}) = O(n_{\epsilon,\delta}).
\end{equation}
To see why the inequality holds, note that $\sqrt{\epsilon}\log\frac{1}{\epsilon} \leq O(1)$ and $\beta \geq \Omega(1)$, hence $\sqrt{\epsilon}N = \sqrt{\epsilon} \cdot \frac{1}{\beta^2} \log\frac{1}{\epsilon} \leq O(1)$. Next, the algorithm invokes \complabel on $S_{\text{in}}$ whose size is upper bounded by $\sqrt{\epsilon} \abs{S} = O(n_{\sqrt{\epsilon},\delta})$ in view of Lemma~\ref{lem:size-inside-interval-restate}. Therefore, by Proposition~\ref{prop:label-comp-complexity-restate}, the label complexity and comparison of this step is $O\big(\frac{1}{\alpha^2} \cdot \log n_{\sqrt{\epsilon},\delta} \cdot \log\log n_{\sqrt{\epsilon},\delta}  \big)$ and $O\big(\frac{1}{\beta^2} \cdot n_{\sqrt{\epsilon},\delta} \cdot \log^2 n_{\sqrt{\epsilon},\delta}\big)$ respectively. These, in conjunction with the label and comparison complexity of the sub-sampling step gives the announced results.
\end{proof}

\subsection{Performance guarantee of Phase 2}

We now switch to the notation in Algorithm~\ref{alg:boost}.

\begin{lemma}[Lemma 4.7 in \citet{awasthi2017efficient}] \label{lem:SI-restate-WI-WC}
With probability $1- \delta/12$, $\bar{W}_I$ and $\bar{W}_C$ both have size $\Theta(n_{\sqrt{\epsilon},\delta})$.
\end{lemma}


\begin{proposition}[Restatement of Prop.~\ref{prop:err-h2}]\label{prop:err-h2-restate}
Assume that $\beta \geq c_0$ for some absolute constant $c_0 \in (0, 1/2]$. In Phase~2, with probability $1-\frac{\delta}{3}$, $\err_{\D_2}(h_2)\leq \frac{1}{2}\sqrt{\epsilon}$. The label complexity is $O\big(\frac{1}{\alpha^2} \cdot \log n_{\sqrt{\epsilon},\delta} \cdot \log{\log n_{\sqrt{\epsilon},\delta}}\big)$, and the comparison complexity is $O\big(n_{\sqrt{\epsilon},\delta} \cdot \log^2 n_{\sqrt{\epsilon},\delta} + n_{\epsilon,\delta} \big)$.
\end{proposition}

\begin{proof}
To see the query complexity in Phase 2, observe that we query the workers when invoking \filter to obtain $S_I$, and when invoking \complabel to obtain $\bar{S_{\text{All}}}$ where $\abs{\bar{S_{\text{All}}}} = \Theta(n_{\sqrt{\epsilon},\delta})$. Therefore, with probability $1-\delta/6$, we obtain the announced query complexity in view of Proposition~\ref{prop:filter-guarantee-restate} and Proposition~\ref{prop:label-comp-complexity-restate}.


It remains to show that $h_2$ achieves error rate $\frac12\sqrt{\epsilon}$ on the target distribution $\D_2$. Let $d(x)$, $d_C(x)$ and $d_I(x)$ be the density functions of $\D$, $\D_C$ and $\D_I$ respectively. Since the error rate of $h_1$ is $\Theta(\sqrt{\epsilon})$ (see Proposition~\ref{prop:err-h1-restate}), we have that for any $x$ with $h_1(x) = h^*(x)$, $d(x) = d_C(x) \cdot (1 - \Theta(\sqrt{\epsilon}))$; for any $x$ with $h_1(x) \neq h^*(x)$, $d(x) = d_I(x) \cdot \Theta(\sqrt{\epsilon})$.

Let $N_C(x)$, $N_I(x)$, $M_C(x)$ and $M_I(x)$  be the number of occurrences of $x$ in the sets $S_C$, $S_I$, $\bar{W_C}$ and $\bar{W_I}$, respectively. Let $d'(x)$ be the density function of the distribution $\D'$ underlying the empirical distribution of $\frac12\bar{W_I} + \frac12\bar{W_C}$.

We condition on Lemma~\ref{lem:SI-restate-WI-WC}, which occurs with probability $1-\delta/12$.

\medskip
\noindent{\bfseries Case 1.} $x$ is such that $h_1(x) = h^*(x)$.  We have
\begin{align*}
d'(x)&= \frac 12 \E\left[ \frac{M_C(x)}{ \abs{\bar{W_C}} } \right] \stackrel{\zeta_1}{\geq} \frac{\E[M_C(x)]}{  \Theta(n_{\sqrt\epsilon,\delta}) } \stackrel{\zeta_2}{\geq} \frac{\E[N_C(x)]}{ \Theta( n_{\sqrt\epsilon,\delta}) } 
= \frac{ \abs{S_C} \cdot d(x) }{ \Theta( n_{\sqrt\epsilon,\delta}) } \\
&= \frac{ \Theta( n_{\sqrt\epsilon,\delta}) \cdot d_C(x) \cdot (1- \Theta(\sqrt{\epsilon})) }{ \Theta( n_{\sqrt\epsilon,\delta}) } 
\geq \Theta( d_C(x)) = \Theta(d_2(x)).
\end{align*}
In the above expression, $\zeta_1$ follows from Lemma~\ref{lem:SI-restate-WI-WC}, $\zeta_2$ holds since $\bar{W_C}$ was obtained in such a way that the majority vote has high probability to be correct while $S_C$ was just drawn from $\D_X$, and the last step simply follows from the fact that $\D_2$ is an equally weighted mixture of $\D_C$ and $\D_I$.

\medskip
\noindent{\bfseries Case 2.} $x$ is such that $h_1(x) \neq h^*(x)$. Similar to the first case, we can show that
\begin{align*}
d'(x) &= \frac 12 \E\left[ \frac{M_I(x)}{\abs{\bar{W_I}}} \right] = \frac{\E[M_I(x)]}{ \Theta( n_{\sqrt\epsilon,\delta}) }
\geq \frac{\E[N_I(x)]}{ \Theta( n_{\sqrt\epsilon,\delta}) }  \geq \frac{ \frac{4c_0}{1+2c_0} \abs{S_2}  d(x)}{ \Theta( n_{\sqrt\epsilon,\delta}) } \\
&
= \frac{ \frac{4c_0}{1+2c_0} \abs{S_2}  d_I(x) \Theta(\sqrt{\epsilon}) }{ \Theta(n_{\sqrt\epsilon,\delta}) } = \frac{\Theta(n_{\sqrt{\epsilon},\delta}) \cdot d_I(x)}{\Theta(n_{\sqrt{\epsilon},\delta})}
=  \Theta(d_I(x)) = \Theta(d_2(x)).
\end{align*}

Now by the super-sampling lemma (Lemma~4.2 in \citet{awasthi2017efficient}), we know that the obtained $h_2$ is such that with probability at least $1-\frac{\delta}{12}$, $\err_{\D_2}(h_2) \leq \frac{1}{2}\sqrt{\epsilon}$. The desired success probability of $1-\delta/3$ follows by considering the union bound of all the events.
\end{proof}

\section{Analysis of Phase 3}\label{sec:app:phase3}

\begin{proposition}[Restatement of Prop.~\ref{prop:err-h3}]\label{prop:err-h3-restate}
In Phase~3, with probability $1-\frac{\delta}{3}$, $\err_{\D_3}(h_3) \leq \frac12\sqrt{\epsilon}$. In addition, the label complexity is $O\big(\frac{1}{\alpha^2}\cdot\log n_{\sqrt{\epsilon},\delta}\cdot\log{\log n_{\sqrt{\epsilon},\delta}}\big)$ and the comparison complexity is $O\big(\frac{1}{\beta^2}\cdot n_{\sqrt{\epsilon},\delta} \cdot\log^2 n_{\sqrt{\epsilon},\delta}\big)$.
\end{proposition}
\begin{proof}
Similar to the proof of Proposition~\ref{prop:err-h1-restate}, in Phase 3, all instances in $S_3$ are correctly labeled with probability at least $1-\frac{\delta}{6}$. The label complexity is $O\big(\frac{1}{\alpha^2}\cdot\log n_{\sqrt{\epsilon},\delta}\cdot\log{\log n_{\sqrt{\epsilon},\delta}}\big)$  and the comparison complexity is $O\big(\frac{1}{\beta^2}\cdot n_{\sqrt{\epsilon},\delta} \cdot\log^2 n_{\sqrt{\epsilon},\delta}\big)$. We condition on this happening. Then with probability $1-\frac{\delta}{6}$, the base PAC learner $\A_\H$ returns a classifier with error rate $\leq \frac12\sqrt{\epsilon}$ in view of Assumption~\ref{as:pac}. By union bound, with probability at least $1-\frac{\delta}{3}$, the error rate of $h_3$ is $\err_{\D_3}(h_3)\leq \frac12\sqrt{\epsilon}$. 
\end{proof}

\section{Proof of Theorem~\ref{thm:main}} \label{sec:app:summary}


\begin{proof}
By Proposition~\ref{prop:err-h1-restate}, Proposition~\ref{prop:err-h2-restate}, and Proposition~\ref{prop:err-h3-restate}, we have that with probability at least $1-\delta$, the  error guarantees $\err_{\D_X}(h_1) \leq \frac12\sqrt{\epsilon}$, $\err_{\D_2}(h_2) \leq \frac12\sqrt{\epsilon}$ and $\err_{\D_3}(h_3) \leq \frac12\sqrt{\epsilon}$ hold simultaneously. Therefore, by Theorem~\ref{thm:boost}, it follows that the classifier $\hat{h} := \maj(h_1, h_2, h_3)$ is such that $\err_{\D}(\hat{h}) \leq \frac34\epsilon \leq \epsilon$ with probability at least $1-\delta$.

Note that each phase of Algorithm~\ref{alg:boost} runs in polynomial time. In particular, the \complabel algorithm runs in polynomial time because its core component is \quicksort which runs in polynomial time. The \filter algorithm runs in polynomial time because the sub-sampling step is efficient and the number of iterations for filtering is $N = \frac{1}{\beta^2}\log\frac{1}{\epsilon}$. Last, the base learner $\A_\H$ is assumed to be efficient.

Finally, we consider the query complexity of the main algorithm. By Propositions \ref{prop:err-h1-restate}, \ref{prop:err-h2-restate} and \ref{prop:err-h3-restate}, the overall label complexity is given by 
\begin{equation}
m_L = O\Big( \frac{1}{\alpha^2} \cdot  \log n_{\sqrt{\epsilon},\delta} \cdot \log {\log n_{\sqrt{\epsilon},\delta}} \Big) = \frac{1}{\alpha^2} \cdot \tilde{O}\Big( \log\frac{d + \frac{1}{\delta}}{\epsilon} \Big),
\end{equation}
and the overall comparison complexity, under the assumption $\beta \geq c_0$, is given by
\begin{equation}
m_C = O\Big( n_{\sqrt{\epsilon},\delta} \cdot \log^2 n_{\sqrt{\epsilon},\delta} + n_{\epsilon,\delta} \Big) = \tilde{O}\Big( \frac{d + (1/\delta)^{\frac{1}{1000}}}{\sqrt{\epsilon}} \Big) + O\Big( \frac{1}{\epsilon}\Big(d \log\frac{1}{\epsilon} + \big(\frac{1}{\delta}\big)^{\frac{1}{1000} } \Big) \Big) = \tilde{O}\Big( \frac{d + (1/\delta)^{\frac{1}{1000}}}{\epsilon} \Big).
\end{equation}
These in allusion to $m_{\epsilon,\delta} = K \cdot \big(\frac{1}{\epsilon}(d \log(1/\epsilon)+\log(1/\delta)\big) \geq \Omega\big(\frac{1}{\epsilon}(d+\log(1/\delta)\big)$ immediately give the overheads as follows:
\begin{equation}\label{eq:labeloverhead}
\labeloverhead = O\Big( \frac{1}{\alpha^2} \cdot \frac{ \log n_{\sqrt{\epsilon},\delta} }{m_{\epsilon,\delta}} \cdot  \log {\log n_{\sqrt{\epsilon},\delta}} \Big) \leq \frac{1}{\alpha^2} \cdot \frac{\epsilon}{d+ \log(1/\delta)} \cdot \tilde{O}\Big( \log\frac{d + \frac{1}{\delta}}{\epsilon} \Big) = \frac{1}{\alpha^2} \cdot \frac{\epsilon}{d} \cdot \tilde{O}\Big( \log\frac{d}{\epsilon} \Big),
\end{equation}
and
\begin{equation}\label{eq:compoverhead}
\compareoverhead = O\Big( \frac{n_{\sqrt{\epsilon},\delta}}{m_{\epsilon, \delta}} \cdot \log^2 n_{\sqrt{\epsilon},\delta} + \frac{n_{\epsilon,\delta}}{m_{\epsilon,\delta}} \Big) \leq O\Big( \sqrt{\epsilon} \cdot \frac{d + (1/\delta)^{\frac{1}{1000}} }{d + \log(1/\delta)} \cdot  \log^2 \big( \frac{d + 1/\delta}{\epsilon} \big) + \frac{d + (1/\delta)^{\frac{1}{1000}} }{d + \log(1/\delta)} \Big).
\end{equation}
Recall that by the definition of $n_{{\epsilon}, \delta}$ and $m_{\epsilon, \delta}$ in Section~\ref{sec:setup}, we have $m_{\epsilon, \delta} = \Theta(n_{\epsilon, \delta}) = \Theta(\frac{d}{\epsilon} \log\frac{1}{\epsilon})$ when $\delta$ is a constant (note that assuming $\delta$ as a constant only simplifies our discussion). In this case, we can see that
\begin{equation*}
\compareoverhead \leq O_{\delta}\Big( \sqrt{\epsilon} \cdot  \log^2 \frac{d}{\epsilon} + 1 \Big).
\end{equation*}
The theorem is proved.
\end{proof}

\begin{remark}\label{rmk:overhead_analysis}
Observe that the denominator we use in the analysis, i.e. $\Omega\big(\frac{d}{\epsilon}\big)$, is the query complexity lower bound of a PAC learner that uses only labels, thus highlighting the saving of labels with access to the comparison queries.
On the other side, if we were to compare our query complexity to the lower bound of a comparison-equipped algorithm in the non-crowdsourcing setting, we can combine Theorem~4.11 and Corollary~4.12 of \cite{kane2017active} to get a (probably loose) lower bound $\Omega(d+\frac{1}{\epsilon})$. With this bound, it is easy to see that now Eq.~\eqref{eq:labeloverhead} and Eq.~\eqref{eq:compoverhead} become
\begin{align*}
\text{(F.3')}\quad \Lambda_L &\leq \frac{1}{\alpha^2} \cdot \frac{1}{d  + 1/\epsilon} \cdot \tilde{O}_{\delta}(\log\frac{d}{\epsilon}),\\
\text{(F.4')}\quad \Lambda_C &\leq O\Big( \frac{\frac{1}{\sqrt{\epsilon}}(d + (1/\delta)^{\frac{1}{1000}})}{d + 1/\epsilon} \cdot \log^2 \frac{d+1/\delta}{\epsilon} + \frac{\frac{1}{\epsilon}(d + (1/\delta)^{\frac{1}{1000}})}{d + 1/\epsilon} \Big) 
= O_{\delta}\Big( \frac{d/\sqrt{\epsilon}}{d+1/\epsilon} \cdot \log^2\frac{d}{\epsilon} + \frac{d/\epsilon}{d+ 1/\epsilon} \Big).
\end{align*}
When $\epsilon \rightarrow 0$, we have $\Lambda_L = o_{\delta}(1)$ and for fixed $d$, $\Lambda_C = O_{\delta}(1)$. Hence, the performance guarantees we highlighted in Remark~\ref{rmk:constant_overhead} still hold for fixed $d$, as we mentioned in Remark~\ref{rmk:alternative_overhead}.
\end{remark}

\section{Useful Lemmas} \label{sec:app:useful-lemmas}


We record some standard results in this section.

\begin{lemma}[High-probability bound for \quicksort]
\label{lem:quicksort}
With probability at least $1-1/m^{c}$ for any constant $c>1$, the following holds. Given an instance set $S$ with $m$ elements, the comparison complexity of sorting all the elements by \textsc{Randomized Quicksort} is $(c +2)m \log_{8/7}m$.
\end{lemma}
\begin{proof}
\textsc{Quicksort} is a recursive algorithm: in each round, it picks a pivot, splits the problem into two subsets, and recursively calls itself on each subset. The program keeps doing this until all the recursive calls contain at most one element. We consider \textsc{Randomized Quicksort} in our algorithm. Note that \textsc{Randomized Quicksort} differs from \textsc{Quicksort} only in the way it picks the pivots: in each round, it picks a random element in set $S$.

Consider a special element $t\in S$. Let $L_i$ be the size of input in the $i$th level of recursion that contains $t$. Obviously $L_0=m$, and we have
\begin{equation*}
\E[L_i|L_{i-1}] < \frac{1}{2} \cdot \frac{3}{4}L_{i-1}+\frac{1}{2}L_{i-1} \leq \frac{7}{8}X_{i-1},
\end{equation*}
because with probability $\frac{1}{2}$, the pivot ranks between $\frac{1}{4}L_{i-1}$ and $\frac{3}{4}L_{i-1}$; and with probability $\frac{1}{2}$, the size of the subset does not shrink significantly.
By the tower rule $\E[X]=\E[\E[X\mid Y]]$, it follows that
\begin{equation*}
\E[L_i] = \E_y[L_i|L_{i-1}=y] \leq \E_{L_{i-1}=y}\bigg[\frac{7}{8}y\bigg] 
= \frac{7}{8}\E\big[L_{i-1}\big] 
\leq \bigg(\frac{7}{8}\bigg)^i \E[L_0] = \bigg(\frac{7}{8}\bigg)^i m.
\end{equation*}
Let $M = (c+2)\cdot\log_{8/7}m$ for some constant $c > 0$. Then, the above inequality gives
\begin{equation*}
\E[L_M] \leq \bigg(\frac{7}{8}\bigg)^M \cdot m \leq \frac{1}{m^{c+2}}\cdot m = \frac{1}{m^{c+1}}.
\end{equation*}
Applying Markov's inequality, we have 
\begin{equation*}
\Pr[L_M \geq 1] \leq \frac{\E[L_M]}{1} \leq \frac{1}{m^{c+1}},
\end{equation*}
which denotes the probability that element $t$ participates in more than $M$ recursive calls. Taking a union bound over all $m$ elements, the probability that any element participates in more than $M$ recursive calls is at most $\(1/m^{c+1}\)\cdot m = 1/m^c$. In other words, with probability at least $1 - 1/m^c$, the total number of pairwise comparisons performed is upper bounded by $m \cdot M = (c+2) m \log_{8/7}m$.
\end{proof}


\begin{lemma}[Probability of Ruin \cite{feller2008introduction}] \label{lem:ruin}
Consider a player who starts with $i$ dollars  against an adversary that has $N$ dollars. The player bets one dollar in each gamble, which he wins with probability $p$. The probability that the player ends up with no money at any point in the game is 
\[\frac{1 - \left( \frac{p}{1-p} \right)^{N} } {1 - \left( \frac{p}{1-p} \right)^{N+i} }.
\]
\end{lemma}

\begin{lemma}[Chernoff bound]\label{lem:chernoff}
Let $Z_1, Z_2, \dots, Z_n$ be $n$ independent random variables that take value in $\{0, 1\}$. Let $Z = \sum_{i=1}^{n} Z_i$. For each $Z_i$, suppose that $\Pr(Z_i =1) \leq \eta$.  Then for any $\alpha \in [0, 1]$
\begin{equation*}
\Pr\( Z \geq  (1+\alpha) \eta n\) \leq e^{-\frac{\alpha^2 \eta n}{3} }.
\end{equation*}
When $\Pr(Z_i =1) \geq \eta$, for any $\alpha \in [0, 1]$
\begin{equation*}
\Pr\( Z \leq  (1-\alpha) \eta n\) \leq e^{-\frac{\alpha^2 \eta n}{2} }.
\end{equation*}
The above two probability inequalities hold when $\eta$ equals exactly $\Pr(Z_i = 1)$.
\end{lemma}

%% file: arxiv_main.bbl
\newcommand{\etalchar}[1]{$^{#1}$}
\begin{thebibliography}{PMM{\etalchar{+}}20}

\bibitem[ABHM17]{awasthi2017efficient}
Pranjal Awasthi, Avrim Blum, Nika Haghtalab, and Yishay Mansour.
\newblock Efficient {PAC} learning from the crowd.
\newblock In {\em Proceedings of the 30th Annual Conference on Learning
  Theory}, pages 127--150, 2017.

\bibitem[ABL17]{awasthi2017power}
Pranjal Awasthi, Maria{-}Florina Balcan, and Philip~M. Long.
\newblock The power of localization for efficiently learning linear separators
  with noise.
\newblock {\em Journal of the {ACM}}, 63(6):50:1--50:27, 2017.

\bibitem[AL87]{angluin1987learning}
Dana Angluin and Philip~D. Laird.
\newblock Learning from noisy examples.
\newblock {\em Machine Learning}, 2(4):343--370, 1987.

\bibitem[AM08]{ailon2008efficient}
Nir Ailon and Mehryar Mohri.
\newblock An efficient reduction of ranking to classification.
\newblock In {\em Proceedings of the 21st Conference on Learning Theory}, pages
  87--98, 2008.

\bibitem[BFKV96]{blum1996polynomial}
Avrim Blum, Alan~M. Frieze, Ravi Kannan, and Santosh~S. Vempala.
\newblock A polynomial-time algorithm for learning noisy linear threshold
  functions.
\newblock In {\em Proceedings of the 37th Annual {IEEE} Symposium on
  Foundations of Computer Science}, pages 330--338, 1996.

\bibitem[BH12]{balcan2012robust}
Maria{-}Florina Balcan and Steve Hanneke.
\newblock Robust interactive learning.
\newblock In {\em Proceedings of the 25th Conference on Learning Theory}, pages
  20.1--20.34, 2012.

\bibitem[CBG{\etalchar{+}}09]{chang2009reading}
Jonathan Chang, Jordan~L. Boyd{-}Graber, Sean Gerrish, Chong Wang, and David~M.
  Blei.
\newblock Reading tea leaves: How humans interpret topic models.
\newblock In {\em Proceedings of the 23rd Annual Conference on Neural
  Information Processing Systems}, pages 288--296, 2009.

\bibitem[CD10]{callison2010speech}
Chris Callison{-}Burch and Mark Dredze.
\newblock Creating speech and language data with amazon's mechanical turk.
\newblock In {\em Proceedings of the 2010 Workshop on Creating Speech and
  Language Data with Amazon's Mechanical Turk}, pages 1--12, 2010.

\bibitem[DDS{\etalchar{+}}09]{deng2009imageNet}
Jia Deng, Wei Dong, Richard Socher, Li{-}Jia Li, Kai Li, and Fei{-}Fei Li.
\newblock {ImageNet}: {A} large-scale hierarchical image database.
\newblock In {\em {IEEE} Conference on Computer Vision and Pattern
  Recognition}, pages 248--255, 2009.

\bibitem[DK17]{doshivelez2017roadmap}
Finale Doshi{-}Velez and Been Kim.
\newblock Towards a rigorous science of interpretable machine learning.
\newblock {\em CoRR}, abs/1702.08608, 2017.

\bibitem[DK20]{diakonikolas2020near}
Ilias Diakonikolas and Daniel~M. Kane.
\newblock Near-optimal statistical query hardness of learning halfspaces with
  {M}assart noise.
\newblock {\em arXiv:2012.09720}, 2020.

\bibitem[DKK{\etalchar{+}}21]{diakonikolas2020polynomial}
Ilias Diakonikolas, Daniel~M. Kane, Vasilis Kontonis, Christos Tzamos, and
  Nikos Zarifis.
\newblock Efficiently learning halfspaces with tsybakov noise.
\newblock In {\em Proceedings of the 53rd Annual {ACM} {SIGACT} Symposium on
  Theory of Computing}, pages 88--101, 2021.

\bibitem[DKTZ20]{diakonikolas2020learning}
Ilias Diakonikolas, Vasilis Kontonis, Christos Tzamos, and Nikos Zarifis.
\newblock Learning halfspaces with {Massart} noise under structured
  distributions.
\newblock In {\em Proceedings of the 33rd Annual Conference on Learning
  Theory}, pages 1486--1513, 2020.

\bibitem[DS09]{dekel2009vox}
Ofer Dekel and Ohad Shamir.
\newblock Vox populi: Collecting high-quality labels from a crowd.
\newblock In {\em Proceedings of the 22nd Conference on Learning Theory}, 2009.

\bibitem[Fel08]{feller2008introduction}
Willliam Feller.
\newblock {\em An introduction to probability theory and its applications,
  volume 2}.
\newblock John Wiley \& Sons, 2008.

\bibitem[FGKP06]{feldman2006new}
Vitaly Feldman, Parikshit Gopalan, Subhash Khot, and Ashok~Kumar Ponnuswami.
\newblock New results for learning noisy parities and halfspaces.
\newblock In {\em Proceedings of the 47th Annual {IEEE} Symposium on
  Foundations of Computer Science}, pages 563--574, 2006.

\bibitem[FH10]{furnkranz2010prefer}
Johannes F{\"{u}}rnkranz and Eyke H{\"{u}}llermeier.
\newblock Preference learning and ranking by pairwise comparison.
\newblock In {\em Preference Learning}, pages 65--82. Springer, 2010.

\bibitem[GR06]{guruswami2006hardness}
Venkatesan Guruswami and Prasad Raghavendra.
\newblock Hardness of learning halfspaces with noise.
\newblock In {\em Proceedings of the 47th Annual {IEEE} Symposium on
  Foundations of Computer Science}, pages 543--552, 2006.

\bibitem[GWKP11]{gomes2011crowd}
Ryan Gomes, Peter Welinder, Andreas Krause, and Pietro Perona.
\newblock Crowdclustering.
\newblock In {\em Proceedings of the 25th Annual Conference on Neural
  Information Processing Systems}, pages 558--566, 2011.

\bibitem[Han19]{hanneke2019website}
Steve Hanneke.
\newblock Proper {PAC} learning {VC} dimension bounds.
\newblock Theoretical Computer Science Stack Exchange, 2019.
\newblock (version: 2019-07-11).

\bibitem[Hau92]{haussler1992decision}
David Haussler.
\newblock Decision theoretic generalizations of the {PAC} model for neural net
  and other learning applications.
\newblock {\em Information and Computation}, 100(1):78--150, 1992.

\bibitem[HJV13]{ho2013adaptive}
Chien{-}Ju Ho, Shahin Jabbari, and Jennifer~Wortman Vaughan.
\newblock Adaptive task assignment for crowdsourced classification.
\newblock In {\em Proceedings of the 30th International Conference on Machine
  Learning}, pages 534--542, 2013.

\bibitem[HKL20]{hopkins2020power}
Max Hopkins, Daniel Kane, and Shachar Lovett.
\newblock The power of comparisons for actively learning linear classifiers.
\newblock In {\em Proceedings of the 34th Annual Conference on Neural
  Information Processing Systems}, pages 6342--6353, 2020.

\bibitem[HKLM20]{hopkins2020noise}
Max Hopkins, Daniel Kane, Shachar Lovett, and Gaurav Mahajan.
\newblock Noise-tolerant, reliable active classification with comparison
  queries.
\newblock In {\em Proceedings of the 33rd Annual Conference on Learning
  Theory}, pages 1957--2006, 2020.

\bibitem[HSRW19]{heckel2019active}
R.~Heckel, N.~B. Shah, K.~Ramchandran, and M.~J. Wainwright.
\newblock Active ranking from pairwise comparisons and when parametric
  assumptions don’t help.
\newblock {\em The Annals of Statistics}, 47(6):3099--3126, 2019.

\bibitem[JN11]{JamiesonN11}
Kevin~G. Jamieson and Robert~D. Nowak.
\newblock Active ranking using pairwise comparisons.
\newblock In {\em Proceedings of the 25th Annual Conference on Neural
  Information Processing Systems}, pages 2240--2248, 2011.

\bibitem[KKMS05]{kalai2005agnostic}
Adam~Tauman Kalai, Adam~R. Klivans, Yishay Mansour, and Rocco~A. Servedio.
\newblock Agnostically learning halfspaces.
\newblock In {\em Proceedings of the 46th Annual {IEEE} Symposium on
  Foundations of Computer Science}, pages 11--20, 2005.

\bibitem[KLMZ17]{kane2017active}
Daniel~M. Kane, Shachar Lovett, Shay Moran, and Jiapeng Zhang.
\newblock Active classification with comparison queries.
\newblock In {\em Proceedings of the 58th Annual {IEEE} Symposium on
  Foundations of Computer Science}, pages 355--366, 2017.

\bibitem[KO16]{khetan2016achieve}
Ashish Khetan and Sewoong Oh.
\newblock Achieving budget-optimality with adaptive schemes in crowdsourcing.
\newblock In {\em Proceedings of the 30th Annual Conference on Neural
  Information Processing Systems}, pages 4844--4852, 2016.

\bibitem[KOS11]{karger2011iterative}
David~R. Karger, Sewoong Oh, and Devavrat Shah.
\newblock Iterative learning for reliable crowdsourcing systems.
\newblock In {\em Proceedings of the 25th Annual Conference on Neural
  Information Processing Systems}, pages 1953--1961, 2011.

\bibitem[KSS92]{kearns1992toward}
Michael~J. Kearns, Robert~E. Schapire, and Linda Sellie.
\newblock Toward efficient agnostic learning.
\newblock In {\em Proceedings of the 5th Annual Conference on Computational
  Learning Theory}, pages 341--352, 1992.

\bibitem[KV94]{kearns1994intro}
Michael~J. Kearns and Umesh~V. Vazirani.
\newblock {\em An Introduction to Computational Learning Theory}.
\newblock {MIT} Press, 1994.

\bibitem[MN06]{massart2006risk}
Pascal Massart and {\'E}lodie N{\'e}d{\'e}lec.
\newblock Risk bounds for statistical learning.
\newblock {\em The Annals of Statistics}, pages 2326--2366, 2006.

\bibitem[MT94]{maass1994fast}
Wolfgang Maass and Gy{\"o}rgy Tur{\'a}n.
\newblock How fast can a threshold gate learn?
\newblock In {\em Proceedings of a workshop on computational learning theory
  and natural learning systems (vol. 1): constraints and prospects}, pages
  381--414, 1994.

\bibitem[PMM{\etalchar{+}}20]{pananjady2017worst}
A.~Pananjady, C.~Mao, V.~Muthukumar, M.~J. Wainwright, and T.~A. Courtade.
\newblock Worst-case vs average-case design for estimation from fixed pairwise
  comparisons.
\newblock {\em The Annals of Statistics}, 48(2):1072--1097, 2020.

\bibitem[PNZ{\etalchar{+}}15]{park2015preference}
Dohyung Park, Joe Neeman, Jin Zhang, Sujay Sanghavi, and Inderjit~S. Dhillon.
\newblock Preference completion: Large-scale collaborative ranking from
  pairwise comparisons.
\newblock In {\em Proceedings of the 32nd International Conference on Machine
  Learning}, pages 1907--1916, 2015.

\bibitem[SBW19]{shah2019feel}
Nihar~B. Shah, Sivaraman Balakrishnan, and Martin~J. Wainwright.
\newblock Feeling the bern: Adaptive estimators for bernoulli probabilities of
  pairwise comparisons.
\newblock {\em {IEEE} Transactions on Information Theory}, 65(8):4854--4874,
  2019.

\bibitem[Sch90]{schapire1990strength}
Robert~E. Schapire.
\newblock The strength of weak learnability.
\newblock {\em Machine Learning}, 5:197--227, 1990.

\bibitem[She21a]{shen2021power}
Jie Shen.
\newblock On the power of localized {P}erceptron for label-optimal learning of
  halfspaces with adversarial noise.
\newblock In {\em Proceedings of the 38th International Conference on Machine
  Learning}, pages 9503--9514, 2021.

\bibitem[She21b]{shen2021sample}
Jie Shen.
\newblock Sample-optimal {PAC} learning of halfspaces with malicious noise.
\newblock In {\em Proceedings of the 38th International Conference on Machine
  Learning}, pages 9515--9524, 2021.

\bibitem[Slo88]{sloan1988types}
Robert~H. Sloan.
\newblock Types of noise in data for concept learning.
\newblock In {\em Proceedings of the First Annual Workshop on Computational
  Learning Theory}, pages 91--96, 1988.

\bibitem[Slo92]{sloan1992Corrigendum}
Robert~H. Sloan.
\newblock Corrigendum to types of noise in data for concept learning.
\newblock In {\em Proceedings of the Fifth Annual {ACM} Conference on
  Computational Learning Theory}, page 450, 1992.

\bibitem[SZ16]{shah2016nooops}
Nihar~B. Shah and Dengyong Zhou.
\newblock No oops, you won't do it again: Mechanisms for self-correction in
  crowdsourcing.
\newblock In {\em Proceedings of the 33nd International Conference on Machine
  Learning}, pages 1--10, 2016.

\bibitem[SZ21]{shen2021attribute}
Jie Shen and Chicheng Zhang.
\newblock Attribute-efficient learning of halfspaces with malicious noise:
  Near-optimal label complexity and noise tolerance.
\newblock In {\em Proceedings of the 32nd International Conference on
  Algorithmic Learning Theory}, pages 1072--1113, 2021.

\bibitem[TLB{\etalchar{+}}11]{tamuz2011adapt}
Omer Tamuz, Ce~Liu, Serge~J. Belongie, Ohad Shamir, and Adam Kalai.
\newblock Adaptively learning the crowd kernel.
\newblock In {\em Proceedings of the 28th International Conference on Machine
  Learning}, pages 673--680, 2011.

\bibitem[Tsy04]{tsybakov2004optimal}
Alexander~B. Tsybakov.
\newblock Optimal aggregation of classifiers in statistical learning.
\newblock {\em The Annals of Statistics}, 32(1):135--166, 2004.

\bibitem[Val84]{valiant1984theory}
Leslie~G. Valiant.
\newblock A theory of the learnable.
\newblock {\em Communications of the {ACM}}, 27(11):1134--1142, 1984.

\bibitem[Vau17]{vaughan2017making}
Jennifer~Wortman Vaughan.
\newblock Making better use of the crowd: How crowdsourcing can advance machine
  learning research.
\newblock {\em Journal of Machine Learning Research}, 18:193:1--193:46, 2017.

\bibitem[WBBP10]{welinder2010advance}
Peter Welinder, Steve Branson, Serge~J. Belongie, and Pietro Perona.
\newblock The multidimensional wisdom of crowds.
\newblock In {\em Proceedings of the 24th Annual Conference on Neural
  Information Processing Systems}, pages 2424--2432, 2010.

\bibitem[XD20]{xu2020simultaneous}
Austin Xu and Mark~A. Davenport.
\newblock Simultaneous preference and metric learning from paired comparisons.
\newblock In {\em Proceedings of the 34th Annual Conference on Neural
  Information Processing Systems}, pages 454--465, 2020.

\bibitem[XZS{\etalchar{+}}17]{xu2017noise}
Yichong Xu, Hongyang Zhang, Aarti Singh, Artur Dubrawski, and Kyle Miller.
\newblock Noise-tolerant interactive learning using pairwise comparisons.
\newblock In {\em Proceedings of the 31st Annual Conference on Neural
  Information Processing Systems}, pages 2431--2440, 2017.

\bibitem[ZL21]{zhang2021improved}
Chicheng Zhang and Yinan Li.
\newblock Improved algorithms for efficient active learning halfspaces with
  {M}assart and {T}sybakov noise.
\newblock In {\em Proceedings of the 34th Annual Conference on Learning
  Theory}, pages 4526--4527, 2021.

\bibitem[ZSA20]{zhang2020efficient}
Chicheng Zhang, Jie Shen, and Pranjal Awasthi.
\newblock Efficient active learning of sparse halfspaces with arbitrary bounded
  noise.
\newblock In {\em Proceedings of the 34th Annual Conference on Neural
  Information Processing Systems}, pages 7184--7197, 2020.

\end{thebibliography}
